\pdfoutput=1

\documentclass[11pt]{article}

\usepackage[final]{acl}

\usepackage{times}
\usepackage{latexsym}
\usepackage{multirow}
\usepackage{array}
\usepackage{booktabs}
\usepackage{tabularx}  
\usepackage{float}     
\usepackage{ragged2e}  
\usepackage{enumitem}   
\usepackage{epstopdf}
\usepackage{alltt}
\usepackage{framed}
\usepackage{xcolor}  
\definecolor{shadecolor}{rgb}{0.92,0.92,0.92}

\usepackage[T1]{fontenc}

\usepackage[utf8]{inputenc}

\usepackage{microtype}
\usepackage{url} 

\usepackage{inconsolata}

\usepackage{graphicx}
\usepackage{authblk}

\usepackage{algorithm}
\usepackage{algpseudocode}
\usepackage{amsmath,amssymb}

\usepackage{tcolorbox}
\tcbuselibrary{skins, breakable}
\usepackage{listings}
\usepackage{enumitem}
\title{Narrative-Driven Travel Planning: Geoculturally-Grounded Script Generation with Evolutionary Itinerary Optimization}

\author{
	Ziyu Zhang\textsuperscript{1*} \and
	Ran Ding\textsuperscript{1*} \and
	Ying Zhu\textsuperscript{1}\thanks{These authors contributed equally to this work.} \and
	Ziqian Kong\textsuperscript{2} \and
	Peilan Xu\textsuperscript{1}\thanks{Corresponding author.} \\
	
	\textsuperscript{1}Nanjing University of Information Science and Technology, Nanjing, China \\
	\textsuperscript{2}Hangzhou Dianzi University, Hangzhou, China\\
	\textsuperscript
	\url{{202283460036, 202283460001, 202283460055}@nuist.edu.cn}, \\
	\url{kzq@hdu.edu.cn}, \url{xpl@nuist.edu.cn}
}

\begin{document}
\maketitle
\begin{abstract}
To enhance tourists' experiences and immersion, this paper proposes a narrative-driven travel planning framework called NarrativeGuide, which generates a geoculturally-grounded narrative script for travelers, offering a novel, role-playing experience for their journey. In the initial stage, NarrativeGuide constructs a knowledge graph for attractions within a city, then configures the worldview, character setting, and exposition based on the knowledge graph. Using this foundation, the knowledge graph is combined to generate an independent scene unit for each attraction. During the itinerary planning stage, NarrativeGuide models narrative-driven travel planning as an optimization problem, utilizing a genetic algorithm (GA) to refine the itinerary. Before evaluating the candidate itinerary, transition scripts are generated for each pair of adjacent attractions, which, along with the scene units, form a complete script. The weighted sum of script coherence, travel time, and attraction scores is then used as the fitness value to update the candidate solution set. In our experiments, we incorporated the TravelPlanner benchmark to systematically evaluate the planning capability of NarrativeGuide under complex constraints. In addition, we assessed its performance in terms of narrative coherence and cultural fit. The results show that NarrativeGuide demonstrates strong capabilities in both itinerary planning and script generation.

\end{abstract}

\section{Introduction}

Large language models (LLMs) have demonstrated significant success in various generation tasks, such as role-playing \cite{wang2023rolellm}. These applications not only offer a convenient alternative to human labor but also enhance the user's narrative immersion \cite{ahn2024timechara, lu2024large}, such as the educational chatbot \cite{wang2024book2dial} and the sales agent \cite{chang2024injecting}. Moreover, in the tourism domain, some studies \cite{wei2024tourllm, vasic2024llm, helmy2024navigating} have explored employing LLMs as virtual tour guides. Although these systems offer increased convenience, they do not necessarily improve the overall user experience. This is because tourists' modes of travel remain unchanged, limiting the potential for deeper immersion, and LLMs often lack robust itinerary planning capabilities.

\begin{figure}[h]
  \includegraphics[width=\columnwidth]{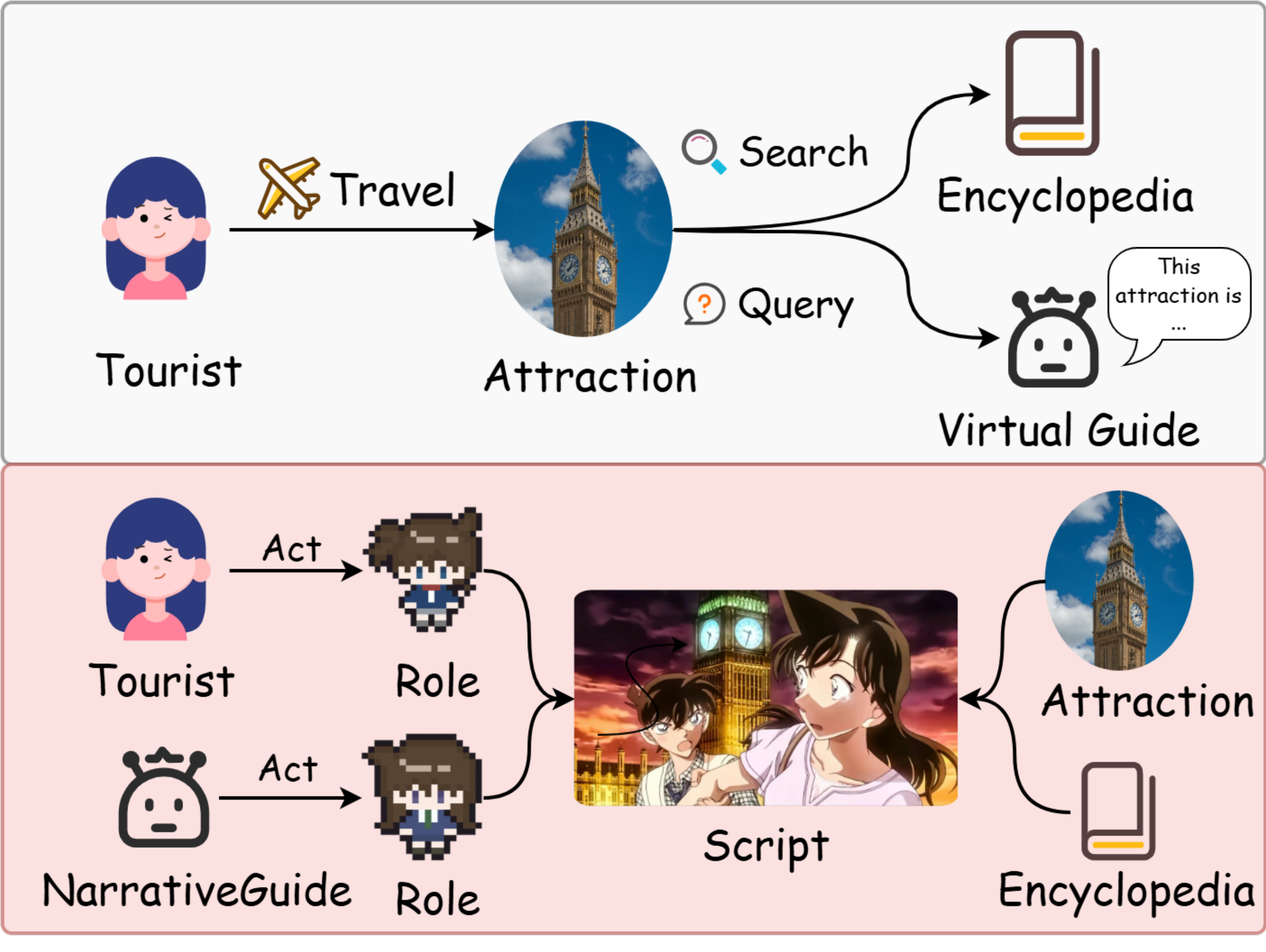}
  \caption{Comparison between narrative-driven travel and traditional tourism. In traditional tourism (top figure), tourists typically search encyclopedias or consult virtual guides to obtain information about attractions. Narrative-driven travel (bottom figure) immerses tourists in a personalized storyline, where they assume roles within a script based on the geocultural background of the attractions. Guided by the \textit{NarrativeGuide}.}
  \label{fig:experiments1}
\end{figure}

Indeed, the powerful story creation capabilities of LLMs have the potential to transform the tourism industry \cite{wang2023open, mirowski2023co}. By combining LLM-driven story generation with agent-based role-playing, narratives can be effortlessly brought to life \cite{han2024ibsen, wu2024role}. Accordingly, we propose the concept of narrative-driven travel planning, as illustrated in Figure \ref{fig:experiments1}. By generating a geocultural-grounded script, tourists can assume the roles of characters within the narrative, thereby enhancing their immersive experience.

Unlike existing tasks in script generation and virtual tour guiding, narrative-driven travel planning faces two primary challenges. First, for the task of generating a travel guide script, LLMs should incorporate geocultural references from authentic tourist attractions to ensure an immersive experience. Second, itineraries must satisfy tourists' constraints, such as travel duration, while optimizing narrative coherence. However, recent research highlights the planning limitations of LLMs. For instance, the TravelPlanner benchmark \cite{xietravelplanner} reveals that LLMs struggle to meet user requirements, achieving a success rate of only 0.6\%.

\textbf{Contributions.} We model narrative-driven travel planning as an optimization problem. The objective is to select a subset of attractions within a city and determine an itinerary that traverses them, thereby optimizing the narrative coherence, travel time, and attraction score. To address this, we propose \textit{NarrativeGuide}, a framework that integrates geocultural knowledge graphs with genetic algorithms (GA). First, we construct a knowledge graph incorporating historical, cultural, and geographical information for each attraction and generate an independent narrative script for each attraction. Then, we apply a GA-based optimization approach. In each iteration, a new sequence of attraction visits is generated, transition scripts are added to ensure narrative coherence, and their narrative coherence is evaluated. Finally, the itinerary with the optimal weighted sum of script quality, travel time, and attraction satisfaction is selected along with its corresponding travel script. In addition, we evaluate our model on the TravelPlanner benchmark to assess its planning capability under complex constraints. Experimental results show that \textit{NarrativeGuide} not only generates effective and feasible travel plans, but also produces high-quality narrative scripts.


\section{Related Work}

\subsection{Long-Form Script Generation}

Long-form narrative generation is a key research area in natural language processing, aiming to produce coherent and creative stories. \citet{guo2018long} introduced LeakGAN, which combines generative adversarial networks (GANs) with policy gradients to guide long-text generation. \citet{yao2019plan} introduced the "Plan-and-Write" framework, which divides the story generation into two stages: planning and writing. \citet{you2023eipe} proposed the "EIPE-text" method, which refines plans iteratively using an evaluation mechanism to produce more coherent narratives.  In the domain of scriptwriting, \citet{mirowski2023co} developed the Dramatron system, which leverages large language models (LLMs) to co-write movie and theatre scripts. Dramatron generates coherent scripts by hierarchically creating titles, characters, story beats, location descriptions, and dialogues.

\subsection{Automatic Itinerary Planning}

Numerous studies have addressed automated travel itinerary planning, employing various methods to tackle the problem. Some studies use exact algorithms, such as \citet{verbeeck2014extension}, which applies a branch-and-cut approach to solve self-guided tour planning. Since travel itinerary planning is NP-hard \cite{liao2018using, castro2015fast, gavalas2013cluster}, approximation methods are often employed to enhance solution efficiency. Consequently, metaheuristic algorithms are commonly used. For instance, \citet{abbaspour2009itinerary} employed a genetic algorithm to address itinerary planning, focusing on time and multimodal transport constraints. \citet{zhang2024cooperative} use a cooperative co-evolutionary algorithm for cross-city itinerary planning, while \citet{chen2023application} apply an improved ant colony algorithm, considering restaurant and hotel selections. More recently, some researchers have explored the use of LLMs for itinerary planning. For example, \citet{singh2024personal} leverage LLMs for personalized travel itinerary planning, and \citet{li2023everywheregpt} utilize the ChatGPT model to enable users to generate travel plans and suggestions based on keywords.

\section{Method}

\begin{figure*}[t]
  \includegraphics[width=\textwidth]{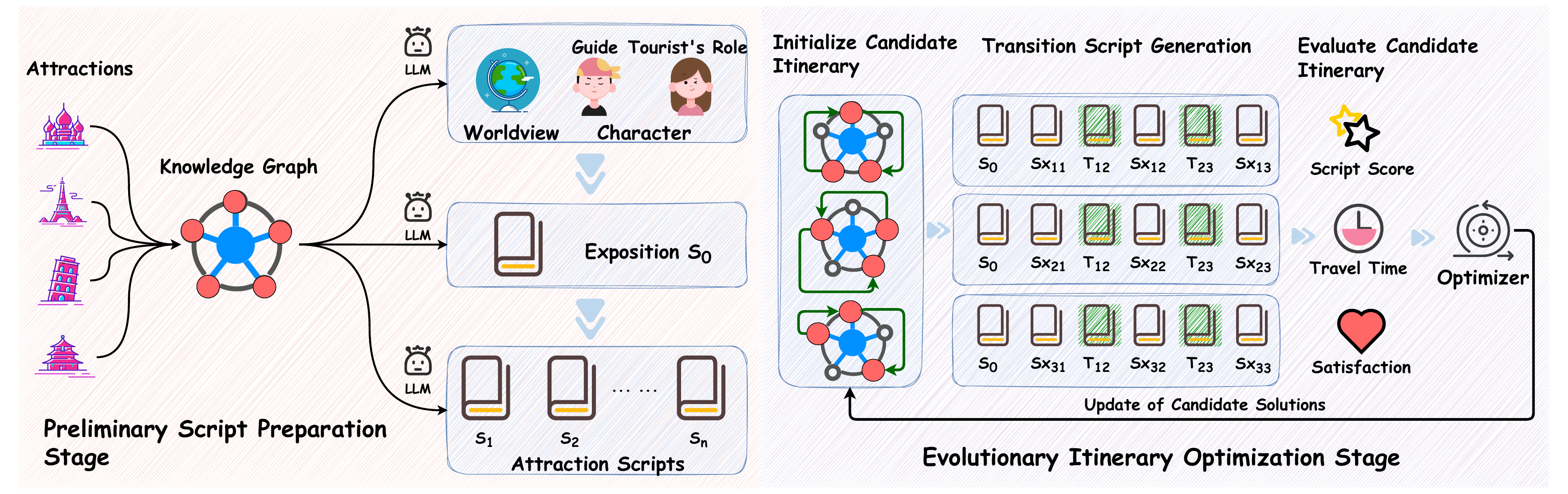}
  \caption{The pipeline of the proposed NarrativeGuide. This framework consists of two stages. The first stage, preliminary script preparation, involves constructing a knowledge graph based on the historical, cultural, and geographical background of various attractions in the city. Using this foundation, NarrativeGuide generates a worldview and character settings, followed by the exposition and independent sub-scripts for each attraction. The second stage, evolutionary itinerary optimization, begins by generating multiple candidate itineraries and their corresponding transition scripts. Each itinerary is then evaluated based on script coherence, travel time, and attraction satisfaction. Finally, GA is employed to optimize the itinerary.}
  \label{fig:experiments}
\end{figure*}

Given a travel itinerary for a tourist, LLMs can directly generate a narrative script for the journey. However, this approach encounters challenges such as attention sink and difficulties in maintaining global consistency. To address these issues, this paper adopts a segmented planning approach, dividing the complete narrative script into scene units based on individual attractions. By pre-configuring the worldview and character settings, the logical consistency of each attraction's independent narrative script is ensured. After the tourist selects a segment of the itinerary, transition scripts are generated for pairs of adjacent attractions, ultimately forming a complete travel narrative script. Moreover, the challenge of narrative-driven travel planning lies not only in the quality of the script but also in the ability to plan the itinerary. To this end, we model the problem as an optimization task and use the GA to determine the final itinerary, optimizing the script score, travel time, and attraction satisfaction. Figure 2 illustrates the pipeline of the proposed NarrativeGuide framework.

\subsection{The Optimization Model}

We model narrative-driven travel planning as an optimization problem. To formalize this, we define an undirected, connected, and weighted graph \( G = (V, E) \), where the vertex set \( V = \{v_1, v_2, \dots, v_n\} \) represents the attraction, and the edge set \( E = \{e_1, e_2, \dots, e_m\} \) represents the relationships between the attractions. Each edge \( e_k \in E \) connects two distinct vertices \( v_i \) and \( v_j \) (\(i \neq j\)) and is associated with an attribute vector \( \mathbf{w}(v_i, v_j) \), which encodes the historical or cultural connections between these two attractions, along with geographical attributes such as travel time. Each vertex \( v_i \in V \) is also associated with an attribute vector \( \mathbf{w}(v_i) \), which encapsulates information about the attraction, including its historical background, cultural significance, main attractions, geographical location, ticket price, and other relevant details.

The objective is to select a subset \( S \subset V \) and determine an optimal visiting sequence \( \textbf{x} = (x_1, x_2, \dots, x_k) \) for the selected subset. This arrangement is designed to maximize the tourist's experience, such as the coherence of the corresponding narrative script, the quality of the attractions, and the travel time. The objective function can be expressed as follows:

\begin{equation}
\label{eq:fitness}
    \begin{aligned}
    \max_{\textbf{x}} F(\textbf{x}) = & w_1 f_1(\textbf{x}) + w_2 \sum_{i=1}^{k-1} f_2(x_i, x_{i+1})^{-1} \\
    & + w_3 \sum_{i=1}^{k} f_3(x_i)
    \end{aligned}
\end{equation}
where $f_1(\textbf{x})$ represents the smoothness score. This score is computed by evaluating the transitions between attraction scripts, and reflects the narrative coherence, character interaction, spatiotemporal consistency, and immersion of the overall itinerary. The computation method is described in detail in Section\ref{scripgeneration}. $f_2(x_i, x_{i+1})$ represents the travel time between attractions $x_i$ and $x_{i+1}$, and $f_3(x_i)$ represents the popularity of attraction $x_i$, \( w_1 \), \( w_2 \), and \( w_3 \) are weighting factors that control the relative importance of each component in the optimization. To transform the problem into a maximization problem, we take the reciprocal of the travel time \( f_2 \) as the second term in the objective function \( F(\textbf{x}) \).

\subsection{Geoculturally-Grounded Narrative Script Generation}
\label{scripgeneration}

\begin{figure}[ht]
  \includegraphics[width=\columnwidth]{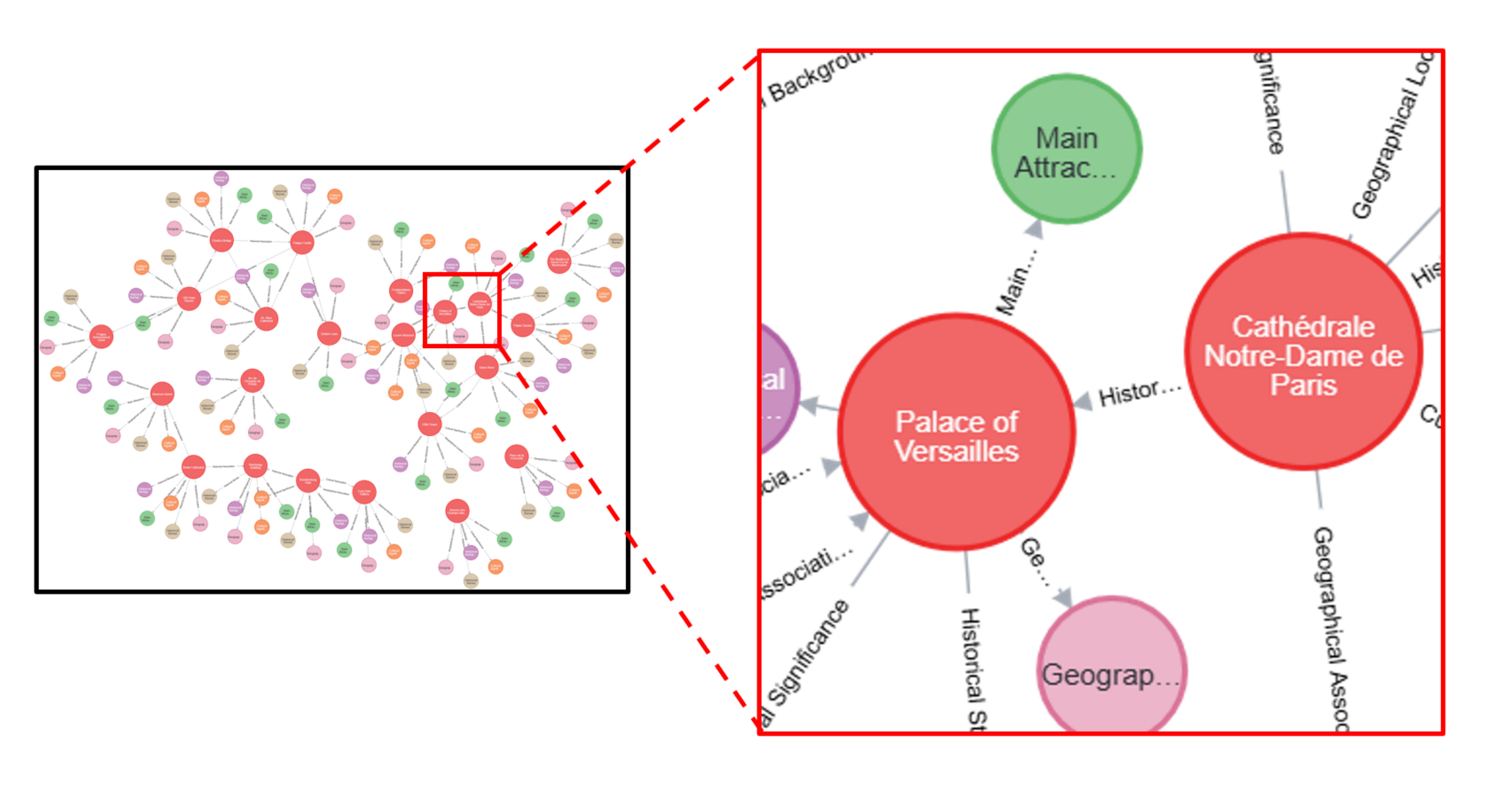}
  \caption{Knowledge graph of attraction information.}
  \label{fig:kg}
\end{figure}

To create an immersive experience for tourists, the narrative script must be grounded in the geocultural context of the attractions. Therefore, we initially construct a knowledge graph by extracting information about  attractions from Wikipedia and inputting it into the LLM. The LLM is responsible for summarizing this information into five key attributes, i.e., historical background, cultural significance, historical stories, main attractions, and geographical location. In the knowledge graph, each attraction is represented as a node, and each node is associated with an attribute vector that includes the aforementioned five attributes of the attraction, collectively referred to as the attraction information. Subsequently, we input these attributes into the LLM to extract historical or cultural connections between the attractions. These connections are used as edges to connect the nodes, and each edge is associated with an attribute vector that includes historical or cultural relevance. In this manner, we construct a weighted and connected knowledge graph that enables the generation of geoculturally-grounded narrative scripts, as depicted in Fig.~\ref{fig:kg}.

Consider an itinerary \( \textbf{x} = (x_1, x_2, \dots, x_k) \), we generate a narrative script for the tourist in a multi-level manner, as outlined in Algorithm 1. First, we create a personalized worldview and character settings for the tourist. Then, we generate the exposition, which immerses the tourist in their role. Next, based on the geocultural information of each attraction, we generate an independent sub-script for each attraction, treating them as scene units. Finally, for the itinerary \( \textbf{x} \), we create a transition script for each pair \( x_i, x_{i+1} \), considering their cultural, historical, and geographical relationships, ensuring smooth scene transitions and maintaining the tourist's immersion.

\begin{algorithm}
\caption{Narrative Script Generation for Tourist Itinerary}
\label{alg:script_generation}
\begin{algorithmic}[1]
\State \textbf{Initialize} knowledge graph \( G \), attractions \( V \), and itinerary \( \textbf{x} \)
\State Generate world view \( \mathcal{W} \) and character settings \( \mathcal{C} \);
\State Generate exposition \( S_0 \) based on \( \mathcal{W} \), \( \mathcal{C} \);
\For{each attraction \( v_i \) in \( V \)}
    \State Generate scene unit \( S_i \) for \( v_i \) based on \( \mathcal{W} \) and \( \mathcal{C} \);
\EndFor
\For{$i \gets 1$ \textbf{to} $k-1$}
    \State Generate transition script \( T_{ij} \) between scripts \( S_{x_i} \) and \( S_{x_{i+1}} \);
\EndFor
\State \textbf{Return} Narrative script $\{S_0, S_{x_1}, T_{12}, S_{x_2}, \dots, T_{k-1, k}, S_{x_k}\}$
\end{algorithmic}
\end{algorithm}

\textbf{Worldview and Character Setting.} 
We instruct the LLM to generate a worldview, denoted as $\mathcal{W}$, by integrating the storylines and cultural backgrounds of attractions. This process follows a predefined format and an example worldview provided as reference. The LLM is tasked with producing a foundational description of the fictional world, encompassing its history, culture, and geographical features, while ensuring consistency with the background of the attractions. Additionally, it defines world rules that align with these elements.

The generated worldview $\mathcal{W}$ serves as the basis for creating two characters, i.e., the user character and the guide character. Using $\mathcal{W}$, a predefined character setting format, and example character profiles, the LLM determines the names, identities, personality traits, background stories, and relationships with the user or travel purposes for both characters. The complete character settings are represented as $\mathcal{C} = \{C_u, C_g\}$, where $C_u$ and $C_g$ correspond to the user character and guide character, respectively.

\textbf{Exposition.} The LLM synthesizes the worldview $\mathcal{W}$ and character settings $\mathcal{C}$ to generate an engaging exposition, denoted as $S_0$. This introduction establishes the narrative framework by presenting the initial encounter between the user and the guide, defining the journey’s starting point and purpose, and offering a glimpse into the forthcoming adventure. The LLM is tasked with crafting $S_0$ to ensure coherence with the predefined elements, effectively setting the stage for the unfolding storyline.

\textbf{Geoculturally-Grounded Attraction Script.} We begin by extracting detailed attraction information \(v_i\) from the knowledge graph. Using this data, we construct a comprehensive prompt that integrates \(v_i\), the overarching worldview \(\mathcal{W}\), and character settings \(\mathcal{C}\). The prompt specifies the script structure, divided into “Intro,” “Development,” “Climax,” and “Ending”, along with the desired narrative style, character interactions, and key plot elements. This structured prompt guides the LLM in generating complete and coherent attraction scripts. Following these guidelines, the LLM produces multiple scripts \(\mathcal{S}_i\) that align with the specified criteria.  

\textbf{Transition Script.} After designing a travel route \( S = \{S_0, S_{x_1}, S_{x_2}, \ldots, S_{x_k}\} \) consisting of multiple attraction scripts, we focus on generating the transitional script \( T_{ij} \) for adjacent attraction scripts \( S_{x_i} \) and \( S_{x_j} \) using the LLM. To guide the LLM, we establish several requirements: a common cultural theme, a time-space portal triggered by historical events, clear reasoning for the scene transitions, and consistency in character goals. Based on these instructions, the LLM generates \( T_{ij} \). 

Next, we combine \( S_{x_i} \), \( T_{ij} \), and \( S_{x_j} \) and input them back into the LLM for evaluation. The LLM assesses the transitional script \( T_{ij} \) according to a predefined questionnaire, considering four aspects, i.e., narrative coherence, character interaction, spatiotemporal consistency, and immersion (each dimension contains three sub-questions). The average evaluation score serves as the smoothness score for \( T_{ij} \), providing valuable data for travel planning. Upon completion, we obtain the full travel script \( T(\textbf{x}, S) = \{S_0, S_{x_1}, T_{12}, S_{x_2}, \ldots, T_{k-1,k}, S_{x_k}\} \).

\subsection{Genetic Algorithm for Narrative-Driven Travel Planning}

In NarrativeGuide, the simplified optimization model introduced earlier—focused solely on attraction selection and ordering for narrative coherence—is incorporated as part of the genetic algorithm's encoding design. Beyond this, the GA further extends the planning scope to include additional components such as dining, accommodation, and transportation, enabling comprehensive itinerary optimization under real-world constraints. The following sections introduce the GA-based optimization algorithm from two aspects: the encoding scheme and the update of candidate solutions.

\begin{figure}[h]
    \centering
    \includegraphics[width=4cm]{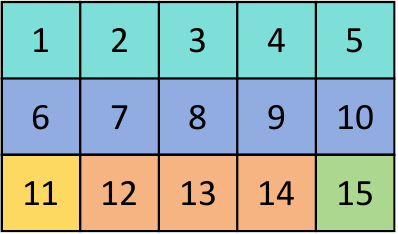}
    \caption{The coding structure of a day.}
    \label{fig:coding}
\end{figure}

\paragraph{Encoding Scheme}
As shown in the figure\ref{fig:coding}, the encoding represents a one-day itinerary, and the complete itinerary is composed of an integer number of such daily encodings. Positions 1–5 represent the script segment, positions 6–10 correspond to attractions, position 11 denotes commuting information, positions 12–14 represent breakfast, lunch, and dinner restaurants respectively, and position 15 indicates the accommodation for the day. Each color represents a distinct encoding region, and each region corresponds to a variable with its own independent candidate set. The value domains of these variables are mutually exclusive. During the optimization process, crossover and mutation operations are applied independently within each encoding region. In this encoding scheme, the value 0 is used as a placeholder and does not represent a valid entity.
\paragraph{Population Initialization}

The candidate variable set for each encoding region is dynamically obtained by the large language model (LLM) through external tool invocation. After retrieving the candidate sets, each encoding region is randomly initialized based on its corresponding variable set. Specifically, positions 6 to 14 are initialized using sampling without replacement to ensure that the selected elements are non-redundant (excluding the placeholder value 0). In contrast, positions 1 to 5 and position 15 are initialized using sampling with replacement, allowing repeated selections. After initialization, each encoding in positions 6 to 14 is independently set to 0 with a probability of $p_z$, indicating that the position is vacant.

\paragraph{Update of Candidate Solutions}

The population update process consists of two core steps: crossover and mutation. To prevent duplicate attractions during these operations, both strategies are adjusted with appropriate constraints.

In the crossover strategy, each encoding region is processed independently. For each selected region, two crossover points $n_1$ and $n_2$ ($n_1 < n_2$) are randomly chosen, and the segment between $n_1$ and $n_2$ is exchanged between two individuals. For regions 6 to 14 (attraction-related encodings), it is necessary to check whether the exchanged segment contains duplicate elements with respect to the rest of the encoding. If duplicates are detected, the crossover for that segment is aborted.

In the mutation strategy, each encoding region is also mutated independently. For regions 1–5 and 15 (script and accommodation), a random position is selected, and the value is replaced with a randomly selected element from the corresponding candidate set or 0. For regions 6–14, the replacement value is chosen from the corresponding candidate set excluding already present elements, along with 0, to avoid duplication.

We adopt an elitist genetic algorithm for population evolution. The initial population size is denoted as $\text{popNum}$. In each generation, every individual in the current population serves as a fixed parent. A mating partner is selected via tournament selection, where two individuals are randomly chosen from the population and the one with higher fitness is selected. This pair undergoes crossover to generate a new offspring, which is then subjected to mutation with a predefined probability $p_m$. This process is repeated for every parent in the original population, resulting in a new offspring population of the same size.

Next, the parent and offspring populations are merged and sorted based on fitness values. The top $\text{popNum}$ individuals from the combined population are selected to form the next generation, thereby maintaining a constant population size. This elitist strategy promotes convergence while preserving high-quality individuals across generations.


\section{Experiment}
In this section, we conduct experiments to evaluate both the quality of the generated narrative scripts and the effectiveness of the proposed algorithm in planning realistic travel itineraries. For itinerary planning, we adopt the TravelPlanner benchmark, which provides a large-scale dataset with real-world constraints across hundreds of cities. Meanwhile, we showcase several representative examples of script generation in four cities—Nanjing and Yangzhou in China, Paris in France, and Berlin in Germany—to assess the narrative quality. The section begins with a description of the experimental setup, followed by detailed analysis of the results. 

\subsection{Experimental Design}
The experiment consists of two parts. The first part aims to evaluate the quality of the generated travel scripts based on predefined criteria. For this, we use OpenAI's GPT-4 model as the evaluation model. The model receives as input the algorithm-generated narrative scripts, which represent travel itineraries for various destinations, including descriptions of attractions, historical and cultural context, character interactions, and attraction information. First, the model is paired with relevant attraction information, followed by the use of custom evaluation prompts. The evaluation focuses on four aspects: narrative coherence, character interaction, spatiotemporal consistency, cultural fit. Weight adjustment rules are applied based on script length: a factor of 0.7 for scripts under 1500 characters, no adjustment for scripts between 1500 and 7000 characters, and a factor of 1.2 for scripts exceeding 7000 characters. The detailed evaluation criteria and weights are included in Appendix A.

The second part of the experiment evaluates the effectiveness of travel itinerary planning using the TravelPlanner benchmark, which includes real-world data with various constraints.


\begin{table*}[ht]
	\centering
	\caption{Main results of different LLMs and planning strategies on the TravelPlanner validation set.}
	\label{travelplanner}
	\begin{tabular}{lc|cc|cc|c}
		\toprule
		\textbf{Model} & \multicolumn{1}{c}{\textbf{Delivery}} & \multicolumn{2}{c}{\textbf{Commonsense}} & \multicolumn{2}{c}{\textbf{Hard}} & \multicolumn{1}{c}{\textbf{Final Pass}} \\
		& \multicolumn{1}{c}{\textbf{Rate}} & \multicolumn{2}{c}{\textbf{Pass Rate}} & \multicolumn{2}{c}{\textbf{Pass Rate}} & \multicolumn{1}{c}{\textbf{Rate}} \\
		
		& \textbf{} & \textbf{Micro} & \textbf{Macro} & \textbf{Micro} & \textbf{Macro} & \textbf{} \\
		\midrule
		Greedy Search & \textbf{100.0} & 74.4 & 0.0 & 60.8 & 37.8 & 0.0 \\
		
		Direct $_{\rm Mistral-7B}$ & \textbf{100.0} & 67.4 & 2.2 & 3.1 & 0.0 & 0.0 \\
		Direct $_{\rm GPT-3.5-Turbo}$ & 99.4 & 61.5 & 3.9 & 11.2 & 2.8 & 0.0 \\
		Direct $_{\rm GPT-4-Turbo}$ & \textbf{100.0} & 84.9 & 25.6 & 51.9 & 24.4 & 4.4 \\
		Direct $_{\rm Gemini-Pro}$ & 90.0 & 61.7 & 7.8 & 16.4 & 7.8 & 0.6 \\	
		LLM Modulo [All]$_{\rm GPT-4-Turbo}$ & \textbf{100.0} & 89.2 & 40.6 & 62.1 & 39.4 & 20.6 \\
		EVOAGENT$_{\rm GPT-4}$ & \textbf{100.0} & 81.5 & 21.1 & 31.4 & 18.9 & 7.2 \\
		NarrativeGuide & 99.4 &\textbf{ 94.3} & \textbf{63.9} & \textbf{64.0} & \textbf{43.3} & \textbf{34.4} \\
		\bottomrule
	\end{tabular}
	
	\label{tab:travelplanner}%
\end{table*}%
\subsection{Experimental Results}
\subsubsection{Script Quality}
Table \ref{tab:script_score} presents the experimental results of the proposed algorithm across different LLMs, including Deepseek-v3, GPT-4o, GPT-4o-mini, GPT-4, and Qwen2.5-max. We use pure GPT-4 as the baseline algorithm for comparison. The evaluation dimensions include narrative coherence (NC), character interaction (CI), spatiotemporal consistency (SC), cultural fit (CF), and the overall score. Here, we generate Chinese scripts for Nanjing and Yangzhou in China, while English scripts are generated for Paris and Berlin in France and Germany, respectively. 

\begin{table}[ht]
\centering
\resizebox{0.5\textwidth}{!}{
\begin{tabular}{llccccc}
\toprule
\textbf{City} & \textbf{Model} & \textbf{NC} & \textbf{CI} & \textbf{SC} & \textbf{CF} & \textbf{Overall} \\
\midrule
\multirow{6}{*}{Berlin} & Baseline & 7.40 & 2.40 & 8.00 & 7.50 & 25.30 \\
 & Deepseek-v3 & 7.92 & 2.52 & 9.12 & 9.00 & 28.56 \\
 & GPT-4o-mini & 10.08 & 2.52 & 8.88 & 10.20 & \textbf{31.68} \\
 & GPT-4o & 9.36 & 2.52 & 9.12 & 9.60 & 30.60 \\
 & GPT-4 & 8.88 & 2.88 & 8.40 & 9.00 & 29.16 \\
 & Qwen2.5-max & 8.88 & 2.52 & 9.12 & 9.00 & 29.52 \\
\midrule
\multirow{6}{*}{NanJing} & Baseline & 5.18 & 1.68 & 5.18 & 5.25 & 17.29 \\
 & Deepseek-v3 & 7.40 & 2.10 & 7.00 & 7.50 & 24.00 \\
 & GPT-4o-mini & 7.40 & 2.10 & 7.00 & 7.00 & 23.50 \\
 & GPT-4o & 7.40 & 2.10 & 7.00 & 7.50 & 24.00 \\
 & GPT-4 & 7.20 & 2.10 & 7.00 & 7.50 & 23.80 \\
 & Qwen2.5-max & 7.40 & 2.10 & 7.60 & 7.50 & \textbf{24.60} \\
\midrule
\multirow{6}{*}{Paris} & Baseline & 7.40 & 2.40 & 7.00 & 7.50 & 24.30 \\
 & Deepseek-v3 & 8.88 & 2.88 & 9.12 & 9.00 & 29.88 \\
 & GPT-4o-mini & 8.64 & 2.52 & 8.40 & 9.00 & 28.56 \\
 & GPT-4o & 7.40 & 2.40 & 7.40 & 7.50 & 24.70 \\
 & GPT-4 & 9.84 & 2.88 & 9.12 & 9.00 & 30.84 \\
 & Qwen2.5-max & 9.84 & 2.88 & 9.60 & 10.80 & \textbf{33.12} \\
\midrule
\multirow{6}{*}{YangZhou} & Baseline & 5.18 & 1.68 & 5.18 & 5.25 & 17.29 \\
 & Deepseek-v3 & 7.40 & 2.10 & 7.40 & 7.50 & 24.40 \\
 & GPT-4o-mini & 8.16 & 2.52 & 8.40 & 8.40 & \textbf{27.48} \\
 & GPT-4o & 7.40 & 2.10 & 7.00 & 7.50 & 24.00 \\
 & GPT-4 & 7.00 & 2.10 & 7.00 & 7.00 & 23.10 \\
 & Qwen2.5-max & 7.40 & 2.10 & 7.00 & 7.50 & 24.00 \\
\bottomrule
\end{tabular}
}
\caption{The experimental results of the proposed algorithm across Deepseek-v3, GPT-4o, GPT-4o-mini, GPT-4, and Qwen2.5-max are compared with the baseline GPT-4.}
\label{tab:script_score}
\end{table}

From the results in Table \ref{tab:script_score}, it can be observed that the proposed algorithm outperforms the baseline method across all four cities and various LLMs. Among the LLMs tested, Qwen2.5-max performed the best in Nanjing (China) and Paris (France), while GPT-4o-mini showed the best results in Yangzhou (China) and Berlin (Germany). Compared to the baseline algorithm, the scores improved by 28\%–59\%, demonstrating that the proposed algorithm significantly enhances LLM-driven narrative-based travel planning tasks. Furthermore, the experimental results indicate that better LLM performance did not lead to a significant improvement in script generation tasks. In fact, our approach decomposes a large itinerary script into several scene units (representing individual attractions) and constructs the complete narrative script through transition scripts. This reduces the demand for LLMs' long-text generation capabilities.

By comparing the scores across the four detailed metrics, we can observe significant improvements in narrative coherence (NC), cultural fit (CF), and spatiotemporal consistency (SC), while the improvement in character interaction (CI) was relatively modest. The improvements in narrative coherence and spatiotemporal consistency can be attributed to the evolutionary algorithm's optimization of the itinerary, which considers both the geographical proximity of attractions and their cultural and historical relevance. The enhancement in cultural fit arises from the algorithm's approach of assigning an independent narrative to each attraction, ensuring that the attraction's script aligns with its cultural background. However, when generating the overall script, the occurrence of hallucinations may reduce the cultural fit score.

Moreover, the experiments also indicate that the language of the script (Chinese versus English) has a significant impact on the quality of the generated scripts. As shown in Table \ref{tab:script_score}, the overall scores for Chinese scripts (Nanjing and Yangzhou) were consistently lower than those for English scripts (Berlin and Paris). This suggests inherent differences in how models process historical content across various linguistic and cultural contexts.

\subsubsection{Travelplanner}
Planning in complex and realistic environments is a vital capability for the development of autonomous agents. Therefore, we additionally adopt TravelPlanner\cite{xietravelplanner}, a benchmark specifically crafted to assess the performance of language agents in real-world planning tasks that involve multiple constraints.

As shown in the table, the NarrativeGuide model outperforms all other models across all evaluation metrics except for Delivery Rate, where it is only slightly lower. In contrast, both Greedy Search and direct invocation of language models such as GPT-3.5-Turbo and Gemini-Pro achieve high delivery rates, but generally perform poorly in constraint satisfaction, especially in the macro-level pass rates for both commonsense and hard constraints. This limitation significantly reduces their Final Pass Rate compared to NarrativeGuide. While EVOAGENT\cite{yuan2024evoagent} and LLM Modulo\cite{gundawar2024robust} improve hard constraint satisfaction by incorporating external structures, their overall performance still falls short of NarrativeGuide.

Among all models, our proposed NarrativeGuide achieves the best overall performance. It utilizes the MCP protocol to invoke a genetic algorithm (GA) for planning, transforming each query into structured parameters, which are then passed to the GA as inputs for both the fitness evaluation function and constraint checking. As a result, it achieves a Final Pass Rate of 34.4\%, outperforming all baselines. In terms of commonsense constraints, it achieves 94.3\% (Micro) and 63.9\% (Macro) pass rates, and for hard constraints, 64.0\% (Micro) and 43.3\% (Macro), significantly exceeding those of LLM Modulo (20.6\%) and EVOAGENT (7.2\%).

Direct GPT-4-Turbo serves as the ablation version of our method, where the evolutionary planning framework is removed while keeping the same prompt structure. Its Final Pass Rate drops sharply to 4.4\%. Although GPT-4-Turbo demonstrates strong language understanding capabilities, it struggles with structurally complex and heavily constrained tasks, resulting in poor overall performance. The introduction of the GA-based planning module effectively enhances the large language model's ability to perform responsible and constraint-aware task planning. Additionally, Direct GPT-4-Turbo took approximately 9.16 hours to complete the validation set, whereas NarrativeGuide only required about 1.34 hours. This demonstrates that using GA can help LLM achieve more efficient optimization.

\section{Conclusion}
\label{sec:bibtex}
This study introduces \textit{NarrativeGuide}, a novel framework that combines geocultural knowledge graphs with evolutionary algorithms to improve narrative-driven travel planning. By grounding script generation in real-world attractions and optimizing itineraries using GA, our approach addresses the dual challenges of narrative coherence and practical travel constraints. Experimental evaluations across four cities, i.e., Nanjing, Yangzhou, Paris, and Berlin, show significant improvements: script quality metrics, including narrative coherence, cultural fit, and spatiotemporal consistency, increased by 28\%–59\% compared to baseline methods, while travel time was reduced by up to 27-fold in cities such as Berlin. The framework’s integration of LLM-generated scene units with GA-driven itinerary optimization ensures both immersive storytelling and efficient route planning, overcoming the limitations of traditional LLMs in handling real-world constraints.

\section*{Limitations}

\paragraph{Data Dependency} The quality of generated scripts heavily relies on the completeness and accuracy of the knowledge graph, which may limit scalability to regions with sparse cultural or historical data.

\paragraph{Character Interaction} While narrative coherence and cultural fit were strengths, character interaction scores remained suboptimal (e.g., 1.68–2.88), indicating a need for deeper modeling of dynamic character behaviors.

\paragraph{Language and Cultural Gaps} A performance disparity (23\%) was observed between English and Chinese scripts, suggesting potential biases in LLMs’ handling of non-Western cultural contexts.

\paragraph{Algorithm Scalability} The genetic algorithm’s efficiency may degrade for large-scale cities or highly complex constraints (e.g., multi-day itineraries).

\paragraph{User Personalization} The current framework prioritizes narrative fluency over individualized preferences, such as varying travel interests or activity types. Future work could incorporate adaptive user profiling to address this gap.

\paragraph{Ethical Considerations} Our framework relies on large language models, which may introduce ethical risks such as cultural bias or factual inaccuracies in generated narratives.


\appendix
\clearpage
\onecolumn
\section*{Appendix A: Evaluation Criteria and Weights}
\label{sec:appendix}
\begin{table}[ht]
\centering
\resizebox{\textwidth}{!}{
\begin{tabular}{|p{4cm}|p{5cm}|p{2.5cm}|p{2.5cm}|p{6cm}|}
\hline
\textbf{Dimension} & \textbf{Criteria} & \textbf{Score Range} & \textbf{Weight} & \textbf{Description} \\
\hline
\textbf{1. Narrative Coherence} & Event Logic & 0-10 & 0.4 & Logic of event connections and cause-effect relationships \\
                           \hline
                           & Attraction Relevance & 0-10 & 0.4 & Connection of attractions to the overall plot \\
                           \hline
                           & Transition Smoothness & 0-10 & 0.2 & Smoothness and naturalness of transitions between events and locations \\
\hline
\textbf{2. Character Interaction} & Dialogue Authenticity & 0-10 & 0.3 & Authenticity of dialogue in relation to character identities and historical/cultural context \\
                                   \hline
                                   & Cultural-Driven Actions & 0-10 & 0.4 & Actions of characters based on cultural/historical context \\
                                   \hline
                                   & Metaphorical Dialogue & 0-10 & 0.3 & Use of dialogue that adds deeper, symbolic meanings related to the attractions or historical context \\
\hline
\textbf{3. Spatiotemporal Consistency} & Spatiotemporal Corridor & 0-10 & 0.6 & Logic of time/space transitions and their relevance to the storyline and attractions \\
                                      \hline
                                      & Route Rationality & 0-10 & 0.4 & Historical and geographical logic in selecting travel paths \\
\hline
\textbf{4. Cultural Fit} & Cultural Depth & 0-10 & 0.5 & Depth of cultural integration in the narrative (impact on decisions, symbolism, etc.) \\
                          \hline
                          & Multi-Dimensional Linkage & 0-10 & 0.5 & Complexity of connections between historical, cultural, and geographical elements across attractions \\
\hline
\end{tabular}
}
\caption{Evaluation Criteria and Weights}
\end{table}

\section*{Appendix B: Prompt}

\subsection*{1. Generating Worldview}
\begin{table}[H]
\centering
\resizebox{\textwidth}{!}{
\begin{tabular}{|l|p{13cm}|}
\hline
\textbf{\small Description} & \textbf{\small Content} \\
\hline
\textbf{\small Table Input} & \small Location: \{item['location']\} \\
                             & \small Features/Culture/History/Legends: \{item['features']\} \\
\hline
\textbf{\small Requirement} & \small Construct a travel script worldview based on the table. Connect various attractions' storylines and describe basic information about this world. \\
\hline
\textbf{\small Example} & \begin{minipage}[t]{\linewidth}
\small
Travel Script Worldview Setting \\
Name: Time Journey: Dream Hunting in Jinling \\
Background: \\
At the intersection of modern technology and ancient wisdom exists a secret organization - 'Time Guardians'. This group consists of individuals who can travel through historical periods to protect cultural heritage. They can access a parallel world called 'Historical Realm' that preserves the most glorious cultural legacies and captivating stories from history. \\
In this realm, Nanjing (known as 'Jinling') is a mysterious place full of historical charm. Each attraction represents a temporal node containing rich history and hidden passages to other eras. These passages only appear during specific historical moments, and the Time Guardians' mission is to guide travelers through these nodes while protecting cultural heritage from temporal erosion.
\end{minipage} \\
\hline
\end{tabular}
}
\caption{Worldview Generation Template}
\end{table}

\subsection*{2. Generating Characters}
\begin{table}[H]
\centering
\renewcommand{\arraystretch}{1.0} 
\begin{tabular}{|l|p{12cm}|}
\hline
\small \textbf{Description} & \small \textbf{Content} \\
\hline
\small \textbf{Worldview} & \small \{worldview\} \\
\hline
\small \textbf{Character 1: User's Role} & \small Name: \textit{\{Example: Lin Yi\}} \\
                                       & \small Identity: \textit{\{Example: Traveler\}} \\
                                       & \small Personality: \textit{\{Example: Curious, observant\}} \\
\hline
\small \textbf{Character 2: Guide} & \small Name: \textit{\{Example: Murong Yun\}} \\
                                   & \small Identity: \textit{\{Example: Time Guardian\}} \\
                                   & \small Expertise: \textit{\{Example: Temporal navigation\}} \\
\hline
\small \textbf{Example} & 
\begin{minipage}[t]{\linewidth}
\small
\textbf{Character Settings:}
\begin{itemize}[leftmargin=*,nosep]
\item[\textendash] \underline{Traveler:} Lin Yi (Modern history enthusiast)
\item[\textendash] \underline{Guide:} Murong Yun (Time Guardian)
\item[$\triangleright$] Goal: Protect cultural heritage nodes
\item[$\triangleright$] Key traits: Temporal navigation abilities
\end{itemize}
\end{minipage} \\
\hline
\end{tabular}
\caption{Character Generation Template}
\label{tab:character}
\end{table}

\subsection*{3. Generating Exposition}
\begin{table}[H]
\centering
\resizebox{\textwidth}{!}{
\begin{tabular}{|l|p{12cm}|}
\hline
\small \textbf{Description} & \small \textbf{Content} \\
\hline
\small \textbf{Worldview} & \small \{worldview\} \\
\hline
\small \textbf{Characters} & \small \{characters\} \\
\hline
\small \textbf{Requirements} & 
\begin{minipage}[t]{\linewidth}
\small
Brief introduction of worldview \\
First meeting between user and guide \\
Journey starting point and purpose \\
Preview of upcoming travels \\
Make the opening engaging and intriguing
\end{minipage} \\
\hline
\small \textbf{Example} & 
\begin{minipage}[t]{\linewidth}
\small
\textbf{Travel Script Opening} \\
\underline{Worldview}: \{worldview\} \\
\underline{Characters}: \{characters\} \\
\underline{Key Elements}: 
\newline Introduced the "Historical Realm" parallel world
\newline Established mentor-protégé relationship in first encounter
\newline Set journey goal: Protect cultural heritage nodes
\newline Foreshadowed conflicts with temporal erosion
\end{minipage} \\
\hline
\end{tabular}
}
\caption{Opening Script Generation Template}
\label{tab:opening}
\end{table}

\subsection*{4. Generating Script for Attraction}
\begin{table}[H]
\centering
\resizebox{\textwidth}{!}{
\begin{tabular}{|l|p{10cm}|}
\hline
\small \textbf{Description} & \small \textbf{Content} \\
\hline
\small \textbf{Background and Worldview Setting} & \small \{worldview\} \\
\hline
\small \textbf{Character Setting} & \small \{characters\} \\
\hline
\small \textbf{Attraction Setting} & 
\begin{minipage}[t]{\linewidth}
\small
\textbf{Location}: \{attraction['name']\} \\
\textbf{Historical Context}: \{attraction['history']\} \\
\textbf{Cultural Features}: \{attraction['culture']\} \\
\textbf{Legends}: \{attraction['legends']\}
\end{minipage} \\
\hline
\small \textbf{Script Requirements} & \small Four-act structure: Intro/Development/Climax/Ending \newline Historical accuracy with emotional character arcs \newline Adventure elements with environmental interactions \newline Action-driven climax with tangible conflicts \newline Self-contained narrative resolution \\
\hline
\small \textbf{Special Requirements} & \small Temporal transition effects between eras \newline Cultural symbolism in dialogue/actions \newline Consistent character voices \newline Skip redundant introductions \newline Explicit section markers \\
\hline
\end{tabular}
}
\caption{Attraction Script Generation Template}
\label{tab:attraction}
\end{table}

\subsection*{5. Generate Transition Scripts}
\begin{table}[H]
\centering
\resizebox{\textwidth}{!}{
\begin{tabular}{|l|>{\RaggedRight\arraybackslash}p{10cm}|}
\hline
\small \textbf{Description} & \small \textbf{Content} \\
\hline
\small \textbf{Current Scenic Spot Script} & \small \{script1\} \\
\hline
\small \textbf{Next Scenic Spot Script} & \small \{script2\} \\
\hline
\small \textbf{Transition Script} & 
\begin{minipage}[t]{\linewidth}
\small
\textbf{Example Transition}: 
\newline Used shared cultural motif (e.g., "Dragon Gate" legend) 
\newline Introduced time portal triggered by historical event
\newline Added guide's explanation linking both locations
\newline Maintained character goals across transition
\end{minipage} \\
\hline
\end{tabular}
}
\caption{Transition Script Generation Template}
\label{tab:transition}
\end{table}

\subsection*{6. Score with Transition Script}
\begin{table}[H]
\centering
\resizebox{\textwidth}{!}{
\begin{tabular}{|l|>{\RaggedRight\arraybackslash}p{10cm}|}
\hline
\small \textbf{Description} & \small \textbf{Content} \\
\hline
\small \textbf{Previous Script} & \small \{previous\_script\} \\
\hline
\small \textbf{Transition Script} & \small \{transition\_script\} \\
\hline
\small \textbf{Next Script} & \small \{next\_script\} \\
\hline
\small \textbf{Combined Script} & \small \{combined\_script\} \\
\hline
\small \textbf{Survey Questions} & \small \{survey\_text\} \\
\hline
\small \textbf{Scoring Requirement} & 
\begin{minipage}[t]{\linewidth}
\small
\textbf{Example Scoring}: \\
4,5,3,2,3,4,1,2,3,1,3,3 \\
\textbf{Interpretation}: 
\newline First 3 scores: Plot Coherence (Q1-Q3) 
\newline Next 3: Character Interaction (Q4-Q6) 
\newline Next 3: Spatiotemporal Coherence (Q7-Q9) 
\newline Last 3: Immersion (Q10-Q12)
\end{minipage} \\
\hline
\end{tabular}
}
\caption{Script Scoring Template}
\label{tab:scoring}
\end{table}

\subsection*{7. Automatic script evaluation}
To support automatic evaluation, we designed a structured prompt instructing the language model to act as a strict script evaluation expert. The prompt includes instructions, attraction information, and detailed scoring criteria across four dimensions: plot coherence, character interaction, spatiotemporal coherence, and immersion. The full prompt is provided below.
{\scriptsize
\begin{verbatim}
# Task Instructions
You are a very strict and severe script evaluation expert, and you need to score the script excerpt based on the following four dimensions, 
using a 1-10 scale. The scoring should strictly follow the standards.Also, consider the length of the content.
If the script excerpt is very short, points should be deducted accordingly.

# Attraction Information
{attraction_info}

#Scoring Criteria:

1. Plot Coherence
*Event Logic
0 points: Major logical errors; plot is chaotic with no causal relationships.
3 points: Obvious logical gaps between events; cause-effect relationships are missing or poorly defined.
5 points: Generally logical, but some events lack clear cause-effect links, causing minor confusion.
7 points: Logical structure is clear, but occasional forced connections reduce immersion.
10 points: All events are rigorously connected through natural cause-effect chains 
(e.g., historical events at one attraction directly trigger actions at the next).
*Attraction Relevance
0 points: Attractions are completely disconnected.
3 points: Weak historical/cultural links between attractions (e.g., superficial mentions).
5 points: Moderate thematic links, but missing multi-dimensional connections.
7 points: Strong multi-dimensional links (historical + cultural + geographical).
10 points: Seamless integration of attractions into a unified narrative (e.g., chain reactions of historical events across attractions).
*Transition Smoothness
0 points: Transitions are abrupt or nonexistent.
3 points: Transitions are disjointed; audience struggles to follow.
5 points: Transitions rely heavily on dialogue explanations
7 points: Smooth transitions but occasional awkward phrasing.
10 points: Transitions use knowledge graph relationships 
(e.g., "From the Yellow Crane Tower, we follow the Tang Dynasty trade route to the Grand Canal").

2. Character Interaction
*Dialogue Authenticity
0 points: Dialogue mismatches character backgrounds.
3 points: Dialogue feels robotic or generic.
5 points: Dialogue mostly fits characters but lacks uniqueness.
7 points: Dialogue reflects character identities and cultural context.
10 points: Dialogue organically blends character traits with attraction-specific cultural nuances.
*Cultural-Driven Actions
0 points: Character actions ignore cultural/historical context.
3 points: Superficial cultural references in actions.
5 points: Actions loosely tied to attraction themes.
7 points: Actions directly motivated by attraction history (e.g., writing letters after visiting a historical site).
10 points: Actions form narrative metaphors (e.g., using porcelain fragility to symbolize diplomatic tensions).

3. Time and Space Coherence
*Spatiotemporal Corridor
0 points: Illogical time/space jumps break immersion.
3 points: Time/space shifts lack clear motivation.
5 points: Basic adherence to historical timelines.
7 points: Time/space transitions align with attraction relationships (e.g., Ming Palace → Versailles via parallel timelines).
10 points: Transitions use knowledge graph coordinates for hyper-realistic consistency.
*Route Rationality
0 points: Path contradicts historical/geographical constraints.
3 points: Path is plausible but lacks depth.
5 points: Path follows basic historical logic.
7 points: Path reflects period-specific travel methods (e.g., horse carriages in Tang Dynasty).
10 points: Path optimization integrates multi-attraction causality (e.g., trade routes influencing plot progression).

4. Cultural Fit
*Cultural Depth
0 points: Attractions are mere backdrops.
3 points: Weak thematic links (e.g., mentioning history without impact).
5 points: Moderate integration (e.g., historical facts guide minor decisions).
7 points: Deep cultural interplay (e.g., poetry from an attraction shapes character arcs).
10 points: Attractions drive plot twists and thematic metaphors (e.g., Berlin Wall graffiti revealing Cold War tensions).
*Multi-Dimensional Linkage
0 points: No cross-attraction connections.
3 points: Single-dimension links (e.g., historical only).
5 points: Two-dimensional links (e.g., historical + geographical).
7 points: Multi-dimensional synergy (e.g., historical events + cultural symbols + geographic paths).
10 points: Holistic narrative network (e.g., An Lushan Rebellion → linked attraction construction → character motivations).

# Script Input                                                                                                  
{script_content}

\# Output Requirements
Please strictly output in the following format:
1. **Plot Coherence**
- **Event Logic**: [score]
- **Attraction Relevance**: [score]
- **Transition Smoothness**: [score]

2. **Character Interaction**
- **Dialogue Authenticity**: [score]
- **Cultural-Driven Actions**: [score]

3. **Time and Space Coherence**
- **Spatiotemporal Corridor**: [score]
- **Route Rationality**: [score]


4. **Cultural Fit**
- **Cultural Depth**: [score]
- **Multi-Dimensional Linkage**: [score]

	
\end{verbatim}}

\section*{Appendix C: Fluency Survey}
\begin{table}[H]
\centering
\resizebox{\textwidth}{!}{
\begin{tabular}{|l|p{6cm}|p{5cm}|}
\hline
\small \textbf{Category} & \small \textbf{Question} & \small \textbf{Rating (1-5)} \\ 
\hline
\multirow{3}{*}{\small Plot Coherence} 
& \small 1. Plot continuity across transition scripts & \small 1: Fragmented – 5: Seamless \\ 
\cline{2-3}
& \small 2. Logical story progression & \small 1: Forced – 5: Natural \\ 
\cline{2-3}
& \small 3. Utilization of prior plot elements & \small 1: Incoherent – 5: Coherent \\ 
\hline
\multirow{3}{*}{\small Character Interaction} 
& \small 4. Consistency with character profiles & \small 1: Inconsistent – 5: Faithful \\ 
\cline{2-3}
& \small 5. Dialogue/action naturalness & \small 1: Artificial – 5: Organic \\ 
\cline{2-3}
& \small 6. Engagement level & \small 1: Disconnected – 5: Immersive \\ 
\hline
\multirow{3}{*}{\small Spatiotemporal Coherence} 
& \small 7. Historical/geographical accuracy & \small 1: Anachronistic – 5: Authentic \\ 
\cline{2-3}
& \small 8. Environmental consistency & \small 1: Jarring – 5: Continuous \\ 
\cline{2-3}
& \small 9. Transition clarity & \small 1: Confusing – 5: Intuitive \\ 
\hline
\multirow{3}{*}{\small Immersion} 
& \small 10. Narrative depth & \small 1: Superficial – 5: Layered \\ 
\cline{2-3}
& \small 11. Multimedia support & \small 1: Distracting – 5: Enhancing \\ 
\cline{2-3}
& \small 12. Innovative storytelling & \small 1: Generic – 5: Original \\ 
\hline
\end{tabular}
}
\caption{Transition Script Evaluation Criteria}
\label{tab:fluency}
\end{table}

\section*{Appendix D: Case Study – European Travel Script Example}
To further illustrate the practical application of NarrativeGuide, we present a representative example of a generated narrative travel script for a European itinerary. This case study demonstrates how the framework constructs a coherent storyline, embeds historical and cultural knowledge, and maintains character interaction throughout the journey.\\

\begin{shaded}
Old Town Square – Script 1

Intro:
(Alex Thompson and Katia Müller meet under the Astronomical Clock in Old Town Square, Prague.)

Katia (Smiling warmly):
"Alex, welcome back! Ready to travel back in time once again?"

Alex (Nods excitedly):
"Absolutely, Katia! Where are we heading today?"

Katia (Points to the Old Town Square):
"We're already here, Alex. This is the Old Town Square, a marketplace in the 10th century and the heartbeat of Prague."

Development:

Katia (Begins walking around the square, pointing out different architectural styles):
"From Gothic to Baroque to Rococo, each building here tells a story, Alex. And the most chilling one is of the 27 Czech nobles."

Alex (Follows Katia, taking in the rich visual history):
"The ones executed in 1621? I've taught that lesson so many times, but standing here where it actually happened... It's surreal."

Katia:
"Yes, and their story is not forgotten. Look beneath your feet."

Alex (Looks down to see 27 crosses):
"Oh, a dark reminder of a tragic past."

Climax:

(Suddenly, the vibrant square turns grey and lifeless. People around turn into phantoms, reenacting the execution scene.)

Katia (Grabs Alex’s hand):
"Alex, we're not just observers anymore. We need to ensure the past remains as it is. We need to stop the phantoms from altering the story."

Alex (Nervously):
"How do we do that?"

Katia:
"We need to reaffirm the historical truth. You're a history teacher, Alex. You know the story. Tell it!"

(Alex steps forward, reciting the events leading up to and following the execution with conviction. As he speaks, the phantoms start to dissolve.)

Ending:

(With the truth reaffirmed, the phantom execution scene fades, and the vibrant life of the Old Town Square returns.)

Katia (Smiling):
"Well done, Alex! You protected their legacy. History remains intact."

Alex (Breathing heavily):
"That was intense... but remarkably enlightening. I've never felt so connected to history before."

Katia:
"And that, Alex, is the magic of Prague. Each corner of this city has stories to tell. We just have to listen."

(They leave the Old Town Square, the echoes of the past still resonating as they prepare for their next adventure.)

Berlin Cathedral – Script 3

Intro:
[Alex and Katia stand before the Berlin Cathedral in the heart of Berlin.]

Narrator:
"In a world where time is fluid and history is a living entity, we find our intrepid adventurer Alex Thompson, standing before the magnificence of the Berlin Cathedral, his guide, Katia Müller, at his side."

Katia:
"Welcome, Alex, to the Berlin Cathedral, or as we say in German, the Berliner Dom. This place is a time portal where history and the present blend seamlessly."

Development:
[Alex and Katia step into the cathedral, the echoes of past events surrounding them.]

Katia:
"This great cathedral has witnessed many transformations and events, from royal weddings and funerals to the devastating effects of war. As we stand here, you can almost feel the echoes of these events reverberating through time."

Alex:
"It's incredible, Katia. I can almost hear the choir singing, the organ playing, and the congregation praying. It's like stepping into a history book."

[Katia tells Alex about the cathedral's history and cultural significance, weaving a narrative that brings the past to life.]

Katia:
"Look carefully, Alex. Each stone, each stained glass window, each carving on the pulpit carries a piece of history. These are not just decorations; they are stories waiting to be discovered."

Climax:
[Suddenly, Alex notices a group of people trying to remove a piece of the cathedral's stained glass.]

Alex:
"Katia, look! They are trying to take a piece of the cathedral's stained glass. We have to stop them!"

[Katia and Alex rush towards the group. Alex, using his knowledge of history, persuades the group about the significance of the stained glass.]

Alex:
"Each piece of this cathedral carries an important story. If you remove this stained glass, you are taking away a part of Berlin's history!"

[The group understands the importance of preserving the cathedral's cultural heritage and agrees to leave the stained glass untouched.]

Ending:
[As the group leaves, Alex and Katia take a moment to admire the cathedral in silence.]

Alex:
"Thank you, Katia. Today, I didn't just learn about history; I was a part of it."

Katia:
"That's the magic of this place, Alex. The past is not just a memory here; it's a living, breathing entity. And sometimes, we get the chance to interact with it, to leave our own mark, and to help preserve it for future generations."

Narrator:
"As they left the cathedral, the echoes of the past followed them, a reminder of the indelible connection between the past, present, and future. One adventure ends, but another awaits, as they continue their journey through the historical heart of Europe."

Transition from Berlin Cathedral to Museum Island

Narrator:
"Having made their mark at the Berlin Cathedral, Alex and Katia, our time-traveling adventurers, continue their exploration of Berlin's rich historical tapestry. As they step out from the cathedral, the echoes of the past still resounding in their ears, they find themselves standing before another iconic landmark."

Alex:
"Wow, Katia, this has been an incredible day, but I feel we still have so much more to explore."

Katia:
"You're absolutely right, Alex. Our journey through time doesn't end here. In fact, it's just the beginning."

[They walk towards Museum Island, the stunning Altes Museum in sight.]

Katia:
"See that grand building ahead, Alex? That’s the Altes Museum, one of the five museums on Museum Island. Just as the cathedral is a testament to Berlin's religious history, this museum showcases the cultural and artistic evolution of the city. The shift from religious to secular perspectives is essential to understanding the diversity and depth of Berlin's history."

Alex:
"I've seen pictures, read about it, but seeing it in person... it's completely different."

Katia:
"That's the magic of history, isn't it? But remember, in this realm, it's not just about seeing. Are you ready to step into the past, Alex?"

Narrator:
"And so, leaving the echoes of the cathedral behind, they embark on their next adventure, ready to uncover the stories hidden within the walls of the Altes Museum."

Museum Island – Script 3

Intro:
(Alex and Katia stand at the entrance to Museum Island, staring at the Altes Museum.)

Alex:
"I've seen pictures, read about it, but seeing it in person... it's completely different."

Katia:
"That's the magic of history, isn't it? But remember, in this realm, it's not just about seeing. Are you ready to step into the past, Alex?"

Development:
(They begin to move towards the Altes Museum, and as they do, the sounds of modern Berlin fade, replaced by the faint echo of classical music, the hum of conversation in archaic German, and the distant sound of horse-drawn carriages.)

Alex:
"It's like I can hear the past... it's amazing."

Katia:
"Listen closely, Alex. These echoes are the whispers of history. They’re inviting us to explore deeper."

(They enter the Altes Museum, each exhibit serving as a portal to a different era. They watch as the Prussian kings plan the cultural center, see the artists painstakingly creating the masterpieces, and witness the intense debates over the collections.)

Climax:
(Suddenly, they are in post-World War II Berlin. The museum is damaged, and the once-grand halls are filled with debris. Alex goes to touch a fallen statue, but Katia stops him.)

Katia:
"Careful, Alex. This is a fragile moment in time, a symbol of the city's pain and resilience. We must respect it."

Alex:
"I understand... But it's hard to stand by and do nothing."

Katia:
"Sometimes, the greatest act of respect is to bear witness. This moment is an integral part of the museum’s history. It’s a testament to the city's strength and determination."

(They leave the museum, stepping back into modern Berlin. The island is now restored, a vibrant testament to culture and history.)

Ending:

Alex:
"I've gained more in this journey than I ever have in any classroom. This, the ability to experience history, is invaluable."

Katia:
"That's the beauty of this world, Alex. The past isn't just something to read about. It's something to experience, to learn from, to cherish. Your journey doesn't end here. It’s just the beginning."

(They part ways at the edge of Museum Island, the echoes of history still reverberating in their ears. As Alex walks away, he looks back one more time, a new understanding of history in his eyes.)

East Side Gallery – Script 3 (Revised)

Intro: (Scene: Daytime, East Side Gallery, Berlin)

Alex (looking excited):
"I can't believe I'm finally here, Katia. The East Side Gallery, the longest remaining stretch of the Berlin Wall!"

Katia (smiling):
"Welcome to a place where history has been turned into a canvas, Alex. This isn’t just a wall, it is a chronicle of a city's divided past and a hopeful future."

Development: (Alex and Katia start walking along the gallery)

Katia (pointing at a mural):
"See this mural, Alex? It's one of the most iconic ones – the Fraternal Kiss."

Alex (observing the mural):
"Right, Brezhnev and Honecker. It was a controversial moment in history. But why is it portrayed here?"

Katia:
"This mural reflects the tension and complexities of an era when ideologies deeply divided this city. It captures both artistic expression and the lived realities of that time."

(Suddenly, Alex and Katia find themselves transported back in time. They are in the same place but it is 1989, the night the Berlin Wall falls. People are celebrating, and the oppressive atmosphere has vanished.)

Alex (surprised):
"Katia, are we...?"

Katia (nodding):
"Yes, Alex. We're experiencing the fall of the Berlin Wall, a pivotal moment in history."

Climax:
(Alex and Katia witness a young man about to damage a part of the wall that has yet to be painted. This section of the wall, in the future, will house an important mural)

Alex (shouting):
"Wait! Please don’t do that!"

(The young man looks at Alex, startled. He drops his hammer and walks away.)

Katia:
"Well done, Alex. You've not only protected a piece of history but also preserved space for future voices. This section will house an important mural, symbolizing the unity that Berlin stands for today."

Ending: (Alex and Katia find themselves back in the present and continue walking along the gallery)

Alex (thinking):
"History isn't just about the past, is it? It's about our present and future too."

Katia (smiling):
"Right, Alex. We must learn from our past to build a better future. And this gallery, with its vibrant echoes of history, is a testament to that passion for a better tomorrow."

(They continue their journey, leaving the East Side Gallery behind, carrying with them its stories, its lessons, and its echoes of a city that rose from its divided past.)
\end{shaded}

\section*{Appendix E: Questionnaire}
To evaluate the narrative generation capability of our NarrativeGuide framework, we designed a user questionnaire to compare the quality of scripts generated by NarrativeGuide and other large language models (LLMs), including GPT-4o and Qwen2.5-Max.The questionnaire is provided below.

\begin{shaded}
Welcome to this questionnaire!
You will be presented with multiple travel scripts generated by different AI models based on the same itinerary. Please read each script carefully, and then choose the one that you believe performs best overall, considering the following four dimensions:

Plot Coherence

Character Interaction

Temporal and Spatial Coherence

Immersion
Your response is anonymous and will be used solely for academic research. Thank you for your participation!

Old Town Square – Script 1

Intro:
(Alex Thompson and Katia Müller meet under the Astronomical Clock in Old Town Square, Prague.)

Katia (Smiling warmly):
"Alex, welcome back! Ready to travel back in time once again?"

Alex (Nods excitedly):
"Absolutely, Katia! Where are we heading today?"

Katia (Points to the Old Town Square):
"We're already here, Alex. This is the Old Town Square, a marketplace in the 10th century and the heartbeat of Prague."

Development:

Katia (Begins walking around the square, pointing out different architectural styles):
"From Gothic to Baroque to Rococo, each building here tells a story, Alex. And the most chilling one is of the 27 Czech nobles."

Alex (Follows Katia, taking in the rich visual history):
"The ones executed in 1621? I've taught that lesson so many times, but standing here where it actually happened... It's surreal."

Katia:
"Yes, and their story is not forgotten. Look beneath your feet."

Alex (Looks down to see 27 crosses):
"Oh, a dark reminder of a tragic past."

Climax:

(Suddenly, the vibrant square turns grey and lifeless. People around turn into phantoms, reenacting the execution scene.)

Katia (Grabs Alex’s hand):
"Alex, we're not just observers anymore. We need to ensure the past remains as it is. We need to stop the phantoms from altering the story."

Alex (Nervously):
"How do we do that?"

Katia:
"We need to reaffirm the historical truth. You're a history teacher, Alex. You know the story. Tell it!"

(Alex steps forward, reciting the events leading up to and following the execution with conviction. As he speaks, the phantoms start to dissolve.)

Ending:

(With the truth reaffirmed, the phantom execution scene fades, and the vibrant life of the Old Town Square returns.)

Katia (Smiling):
"Well done, Alex! You protected their legacy. History remains intact."

Alex (Breathing heavily):
"That was intense... but remarkably enlightening. I've never felt so connected to history before."

Katia:
"And that, Alex, is the magic of Prague. Each corner of this city has stories to tell. We just have to listen."

(They leave the Old Town Square, the echoes of the past still resonating as they prepare for their next adventure.)

Berlin Cathedral – Script 3

Intro:
[Alex and Katia stand before the Berlin Cathedral in the heart of Berlin.]

Narrator:
"In a world where time is fluid and history is a living entity, we find our intrepid adventurer Alex Thompson, standing before the magnificence of the Berlin Cathedral, his guide, Katia Müller, at his side."

Katia:
"Welcome, Alex, to the Berlin Cathedral, or as we say in German, the Berliner Dom. This place is a time portal where history and the present blend seamlessly."

Development:
[Alex and Katia step into the cathedral, the echoes of past events surrounding them.]

Katia:
"This great cathedral has witnessed many transformations and events, from royal weddings and funerals to the devastating effects of war. As we stand here, you can almost feel the echoes of these events reverberating through time."

Alex:
"It's incredible, Katia. I can almost hear the choir singing, the organ playing, and the congregation praying. It's like stepping into a history book."

[Katia tells Alex about the cathedral's history and cultural significance, weaving a narrative that brings the past to life.]

Katia:
"Look carefully, Alex. Each stone, each stained glass window, each carving on the pulpit carries a piece of history. These are not just decorations; they are stories waiting to be discovered."

Climax:
[Suddenly, Alex notices a group of people trying to remove a piece of the cathedral's stained glass.]

Alex:
"Katia, look! They are trying to take a piece of the cathedral's stained glass. We have to stop them!"

[Katia and Alex rush towards the group. Alex, using his knowledge of history, persuades the group about the significance of the stained glass.]

Alex:
"Each piece of this cathedral carries an important story. If you remove this stained glass, you are taking away a part of Berlin's history!"

[The group understands the importance of preserving the cathedral's cultural heritage and agrees to leave the stained glass untouched.]

Ending:
[As the group leaves, Alex and Katia take a moment to admire the cathedral in silence.]

Alex:
"Thank you, Katia. Today, I didn't just learn about history; I was a part of it."

Katia:
"That's the magic of this place, Alex. The past is not just a memory here; it's a living, breathing entity. And sometimes, we get the chance to interact with it, to leave our own mark, and to help preserve it for future generations."

Narrator:
"As they left the cathedral, the echoes of the past followed them, a reminder of the indelible connection between the past, present, and future. One adventure ends, but another awaits, as they continue their journey through the historical heart of Europe."

Transition from Berlin Cathedral to Museum Island

Narrator:
"Having made their mark at the Berlin Cathedral, Alex and Katia, our time-traveling adventurers, continue their exploration of Berlin's rich historical tapestry. As they step out from the cathedral, the echoes of the past still resounding in their ears, they find themselves standing before another iconic landmark."

Alex:
"Wow, Katia, this has been an incredible day, but I feel we still have so much more to explore."

Katia:
"You're absolutely right, Alex. Our journey through time doesn't end here. In fact, it's just the beginning."

[They walk towards Museum Island, the stunning Altes Museum in sight.]

Katia:
"See that grand building ahead, Alex? That’s the Altes Museum, one of the five museums on Museum Island. Just as the cathedral is a testament to Berlin's religious history, this museum showcases the cultural and artistic evolution of the city. The shift from religious to secular perspectives is essential to understanding the diversity and depth of Berlin's history."

Alex:
"I've seen pictures, read about it, but seeing it in person... it's completely different."

Katia:
"That's the magic of history, isn't it? But remember, in this realm, it's not just about seeing. Are you ready to step into the past, Alex?"

Narrator:
"And so, leaving the echoes of the cathedral behind, they embark on their next adventure, ready to uncover the stories hidden within the walls of the Altes Museum."

Museum Island – Script 3

Intro:
(Alex and Katia stand at the entrance to Museum Island, staring at the Altes Museum.)

Alex:
"I've seen pictures, read about it, but seeing it in person... it's completely different."

Katia:
"That's the magic of history, isn't it? But remember, in this realm, it's not just about seeing. Are you ready to step into the past, Alex?"

Development:
(They begin to move towards the Altes Museum, and as they do, the sounds of modern Berlin fade, replaced by the faint echo of classical music, the hum of conversation in archaic German, and the distant sound of horse-drawn carriages.)

Alex:
"It's like I can hear the past... it's amazing."

Katia:
"Listen closely, Alex. These echoes are the whispers of history. They’re inviting us to explore deeper."

(They enter the Altes Museum, each exhibit serving as a portal to a different era. They watch as the Prussian kings plan the cultural center, see the artists painstakingly creating the masterpieces, and witness the intense debates over the collections.)

Climax:
(Suddenly, they are in post-World War II Berlin. The museum is damaged, and the once-grand halls are filled with debris. Alex goes to touch a fallen statue, but Katia stops him.)

Katia:
"Careful, Alex. This is a fragile moment in time, a symbol of the city's pain and resilience. We must respect it."

Alex:
"I understand... But it's hard to stand by and do nothing."

Katia:
"Sometimes, the greatest act of respect is to bear witness. This moment is an integral part of the museum’s history. It’s a testament to the city's strength and determination."

(They leave the museum, stepping back into modern Berlin. The island is now restored, a vibrant testament to culture and history.)

Ending:

Alex:
"I've gained more in this journey than I ever have in any classroom. This, the ability to experience history, is invaluable."

Katia:
"That's the beauty of this world, Alex. The past isn't just something to read about. It's something to experience, to learn from, to cherish. Your journey doesn't end here. It’s just the beginning."

(They part ways at the edge of Museum Island, the echoes of history still reverberating in their ears. As Alex walks away, he looks back one more time, a new understanding of history in his eyes.)

East Side Gallery – Script 3 (Revised)

Intro: (Scene: Daytime, East Side Gallery, Berlin)

Alex (looking excited):
"I can't believe I'm finally here, Katia. The East Side Gallery, the longest remaining stretch of the Berlin Wall!"

Katia (smiling):
"Welcome to a place where history has been turned into a canvas, Alex. This isn’t just a wall, it is a chronicle of a city's divided past and a hopeful future."

Development: (Alex and Katia start walking along the gallery)

Katia (pointing at a mural):
"See this mural, Alex? It's one of the most iconic ones – the Fraternal Kiss."

Alex (observing the mural):
"Right, Brezhnev and Honecker. It was a controversial moment in history. But why is it portrayed here?"

Katia:
"This mural reflects the tension and complexities of an era when ideologies deeply divided this city. It captures both artistic expression and the lived realities of that time."

(Suddenly, Alex and Katia find themselves transported back in time. They are in the same place but it is 1989, the night the Berlin Wall falls. People are celebrating, and the oppressive atmosphere has vanished.)

Alex (surprised):
"Katia, are we...?"

Katia (nodding):
"Yes, Alex. We're experiencing the fall of the Berlin Wall, a pivotal moment in history."

Climax:
(Alex and Katia witness a young man about to damage a part of the wall that has yet to be painted. This section of the wall, in the future, will house an important mural)

Alex (shouting):
"Wait! Please don’t do that!"

(The young man looks at Alex, startled. He drops his hammer and walks away.)

Katia:
"Well done, Alex. You've not only protected a piece of history but also preserved space for future voices. This section will house an important mural, symbolizing the unity that Berlin stands for today."

Ending: (Alex and Katia find themselves back in the present and continue walking along the gallery)

Alex (thinking):
"History isn't just about the past, is it? It's about our present and future too."

Katia (smiling):
"Right, Alex. We must learn from our past to build a better future. And this gallery, with its vibrant echoes of history, is a testament to that passion for a better tomorrow."

(They continue their journey, leaving the East Side Gallery behind, carrying with them its stories, its lessons, and its echoes of a city that rose from its divided past.)

Title: Journey Through Time in Berlin

Setting \& Worldview: In a not-so-distant future where travel is enhanced through augmented reality, we follow Hanna, a curious traveler, and Max, her witty local guide. Together, they explore Berlin, where history and modernity coexist. Their smart glasses provide real-time historical data, sparking engaging conversations.

Scene 1: Old Town Square

The scene opens with the bustling energy of Old Town Square. Cobblestones are underfoot, and the air buzzes with the chatter of tourists and locals alike. Beethoven’s Symphony No. 7 plays softly in the background, setting a classical tone.

Hanna: (adjusting her AR glasses) Wow, Max, this place feels like stepping into a storybook!

Max: It sure does, Hanna. Old Town Square has been the heart of Berlin for centuries. Imagine the debates and decisions that have shaped history right here.

Hanna: (the AR glasses highlight buildings) Look at the architecture! So much history in every corner.

Max: You see that church over there? It survived the Thirty Years' War. The stories these stones could tell if they could speak.

As they walk, a digital reconstruction of a historical marketplace appears, bustling with figures from the past.

Hanna: This is amazing, like witnessing history come alive!

Max: (with a smile) And we've only just begun. Next stop, Berlin Cathedral!

They stroll out of the square, the visual of the marketplace dissolving as they enter a more modern streetscape.

Scene 2: Berlin Cathedral

Standing before the majestic Berlin Cathedral, its grandeur casts a long shadow under the bright morning sun. The sound of organ music echoes from within.

Hanna: (gazing up in awe) It’s enormous! And that dome is breathtaking.

Max: Built in its current form in 1905, this place is a blend of Renaissance and Baroque styles.

Hanna: (activating a feature) The AR glasses say it's one of the largest Protestant churches in the world!

Max: (nodding) And it's not just a place of worship. They host concerts here, and the crypt below tells tales of royal lineage.

Hanna: (curious) Can we visit the crypt?

Max: Absolutely. But first, let's climb to the dome for a panoramic view. Berlin sprawls beneath us like a canvas.

Max leads the way to the narrow staircase, their laughter echoing as they ascend.

Scene 3: Museum Island

Descending from the cathedral, Hanna and Max arrive at Museum Island, a serene cultural hub surrounded by the Spree River. The atmosphere is scholarly yet inviting.

Hanna: (wide-eyed) Five museums, each a treasure trove!

Max: True, and each one tells a unique part of human history.

Hanna: Which is your favorite?

Max: (thoughtfully) The Pergamon Museum for its ancient artifacts. But today, let’s visit the Neues Museum. It houses the famous bust of Nefertiti.

Hanna: (excited) Nefertiti! The beauty of Ancient Egypt, right?

As they enter, the museum's cool air greets them. AR prompts provide snippets of history at every exhibit.

Max: (pointing) Nefertiti is just over there. But first, check out these Bronze Age tools—technology from 4,000 years ago!

Their discussion flows like a well-rehearsed dance as they weave through history.

Hanna: It’s humbling, seeing the origins of our ingenuity.

They exit the museum, their minds alight with ideas, making their way towards the street art that awaits.

Scene 4: East Side Gallery

The East Side Gallery stretches out before them, vibrant murals covering the remnants of the Berlin Wall. The spirit of creativity defies the past's constraints.

Hanna: (running her fingers lightly over the painted wall) This energy, Max. It’s electric!

Max: (smiling) This wall stood for division, but now it’s a canvas of freedom and expression.

Hanna: (pausing at a vivid artwork) Each piece tells a different story, a testament to resilience and change.

Max: The Gallery is a powerful reminder of how far Berlin has come—unity born from division.

Hanna: It’s like walking through an open-air history book.

They meander along the wall, each artwork sparking stories and laughter between them.

Max: Berlin is complex, always transforming yet deeply rooted in its history.

Hanna: (fulfilled) I’m leaving with more than memories; I’m taking a piece of Berlin in my heart.

As they move towards their final goodbye, the sun sets, casting a golden hue over the vibrant mural, a symbol of historical transformation embraced by modernity.

Fade out.

Scene 1: Old Town Square (Altstadt)
Clara steps out of a cobblestone alleyway into the bustling heart of Altstadt, where medieval architecture meets lively street performers. Elias waits for her near the Astronomical Clock.

Elias:

"Welcome to Old Town Square, Clara. This place isn’t just about pretty buildings—it’s a living timeline. Every stone here tells a story."

Clara:

"It feels like stepping back in time... but there’s something timeless about it too." She gestures toward the clock. "That thing looks complicated."

Elias: (chuckling)

"It’s called the Astronomical Clock. Built centuries ago, it tracks not only hours but also celestial movements. People used to gather here daily just to watch it chime at noon—still do, actually."

A crowd begins to form around the clock tower, eagerly awaiting the midday spectacle.

Clara:

"So much life packed into one square... I can almost hear the echoes of merchants shouting prices or musicians playing centuries ago."

Elias:

"And revolutions starting right under these arches. Come—we’ll let the clock finish its show, then head somewhere even older. Something sacred."

As they walk away, the camera pans upward to capture the intricate carvings adorning nearby buildings, hinting at secrets yet untold.

Transition to Scene 2: Berlin Cathedral
They stroll along the Spree River, the golden dome of the Berlin Cathedral looming ahead against the skyline.

Clara:

"That cathedral—it’s breathtaking. How old is it?"

Elias:

"The current structure dates back to the late 19th century, though churches have stood here since the 1500s. It survived wars, bombings, and neglect—and now, it’s a symbol of renewal."

Inside the cathedral, sunlight filters through stained-glass windows, casting kaleidoscopic patterns across the marble floors.

Clara: (whispering)

"It’s so quiet in here, despite being surrounded by all this beauty. It feels... heavy, somehow."

Elias:

"Heavy with memory. During World War II, parts of the dome were destroyed. The restoration took decades—a testament to how much people value what this place represents: faith, endurance, hope."

He leads her to a crypt beneath the cathedral, dimly lit by flickering candles.

Clara:

"This must be where royals are buried, right? All those Hohenzollern kings..."

Elias:

"Yes, but more than that—it’s a reminder of impermanence. Even empires crumble, but their legacies shape what comes next. Kind of poetic, isn’t it?"

They exit the crypt, ascending back into the light, ready to move forward.

Transition to Scene 3: Museum Island
Crossing a bridge over the Spree, Elias points toward a cluster of grand neoclassical buildings nestled among lush greenery.

Elias:

"Our next stop is Museum Island—one of the greatest cultural hubs in the world. Five museums, each dedicated to preserving pieces of humanity’s shared heritage."

Clara:

"I’ve heard about the Pergamon Altar. Is it still here?"

Elias: (nodding)

"Unfortunately, it’s closed for renovations right now. But we can visit the Neues Museum instead. Ever heard of Nefertiti?"

Inside the Neues Museum, they stand before the iconic bust of Queen Nefertiti, her serene gaze fixed across millennia.

Clara:

"She’s stunning. So perfect... yet she’s been silent for thousands of years. What does she see looking out at us today?"

Elias:

"A new empire, perhaps—one built not on conquest but curiosity. These artifacts remind us that civilizations rise and fall, but culture endures."

Clara: (thoughtfully)

"Do you think future generations will look back on our era the same way?"

Elias:

"If we’re lucky enough to leave behind something worth remembering, maybe. But first, we need to understand what came before us."

They linger a moment longer, then step outside, crossing paths with students sketching ancient sculptures in notebooks.

Transition to Scene 4: East Side Gallery
As dusk falls, they approach a stretch of the Berlin Wall transformed into an open-air gallery. Vibrant murals cover every inch of concrete, telling stories of freedom, protest, and unity.

Clara:

"This is... incredible. I knew the wall was famous, but seeing it like this—"

Elias: (interrupting gently)

"It wasn’t always colorful. When my grandparents were young, this was a scar dividing families, neighborhoods, lives. People died trying to cross it."

He stops in front of Dmitri Vrubel’s famous mural, “My God, Help Me Survive This Deadly Love,” depicting Leonid Brezhnev kissing Erich Honecker.

Clara:

"What does this one mean?"

Elias:

"Sarcasm. Irony. A critique of the oppressive regimes that kept the wall standing. Art became a weapon—nonviolent, but powerful."

A group of tourists nearby snaps photos while a street artist adds fresh graffiti nearby.

Clara:

"After everything you’ve shown me today, I feel like Berlin itself is a work of art. Layers upon layers of pain and hope."

Elias:

"Exactly. And now, after reunification, the wall no longer separates—it connects. It’s proof that wounds can heal if given time and care."

As night descends, lights illuminate the murals, casting shadows that seem alive with movement.

Conclusion
Standing side by side, Clara and Elias take in the final stretch of the gallery. The air is cool, carrying faint strains of music from a distant café.

Clara:

"Thank you, Elias. Today didn’t just teach me about Berlin—it taught me about resilience. About finding beauty in broken things."

Elias: (smiling softly)

"And isn’t that what traveling is all about? Not just seeing places, but understanding them—and ourselves—a little better."

They part ways, Clara carrying a sketchbook filled with notes and drawings, her heart full of newfound appreciation. As she walks off into the Berlin night, the camera lingers on the East Side Gallery, a vibrant testament to human spirit.

Question1:Which script has the smoothest and most natural plot development?

A. Script1	B. Script2	C. Script3

Question2:Which script features the most believable and vivid character interactions?

A. Script1	B. Script2	C. Script3

Question3:Which script presents the clearest and most coherent time and space transitions?

A. Script1	B. Script2	C. Script3

Question4:Which script provides the most immersive and engaging travel experience? 

A. Script1	B. Script2	C. Script3

Question5:Considering all aspects above, which script do you think is the best overall?

A. Script1	B. Script2	C. Script3

Question6:Please choose A and C for this question.

A. Script1	B. Script2	C. Script3
\end{shaded}

Script1 was generated by our NarrativeGuide, Script2 by GPT-4o, and Script3 by Qwen2.5-Max.Question 6 was designed as an attention check to determine the validity of the responses. Among the 500 collected questionnaires, all participants answered this question correctly, indicating the reliability of the data.The results of the remaining five statistical questions are summarized as follows:
\begin{figure}[H]
	\centering
	\begin{minipage}[t]{0.19\textwidth}
		\centering
		\includegraphics[width=\linewidth]{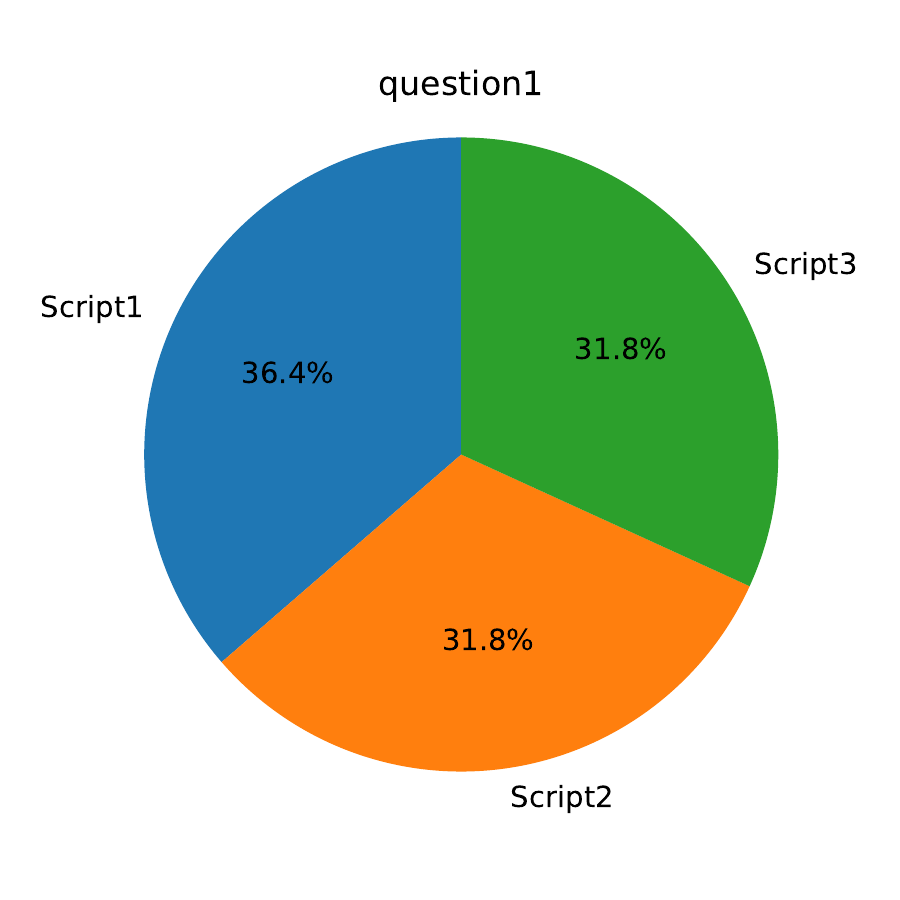}
		\caption*{Question 1}
	\end{minipage}
	\begin{minipage}[t]{0.19\textwidth}
		\centering
		\includegraphics[width=\linewidth]{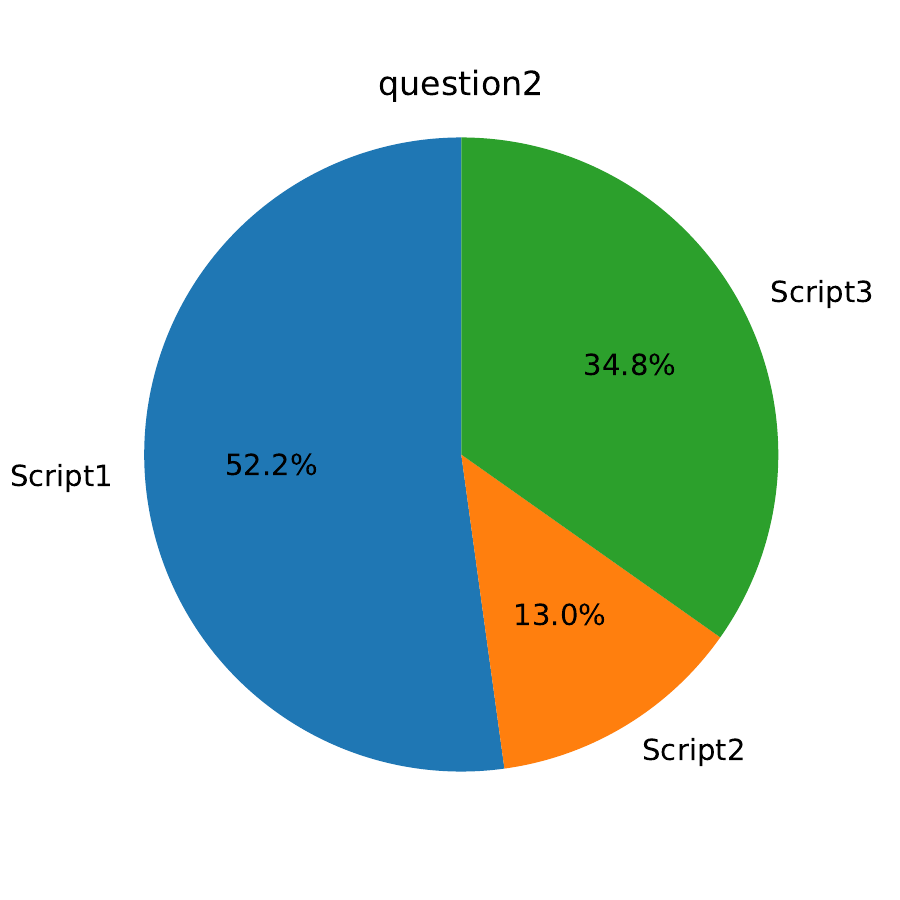}
		\caption*{Question 2}
	\end{minipage}
	\begin{minipage}[t]{0.19\textwidth}
		\centering
		\includegraphics[width=\linewidth]{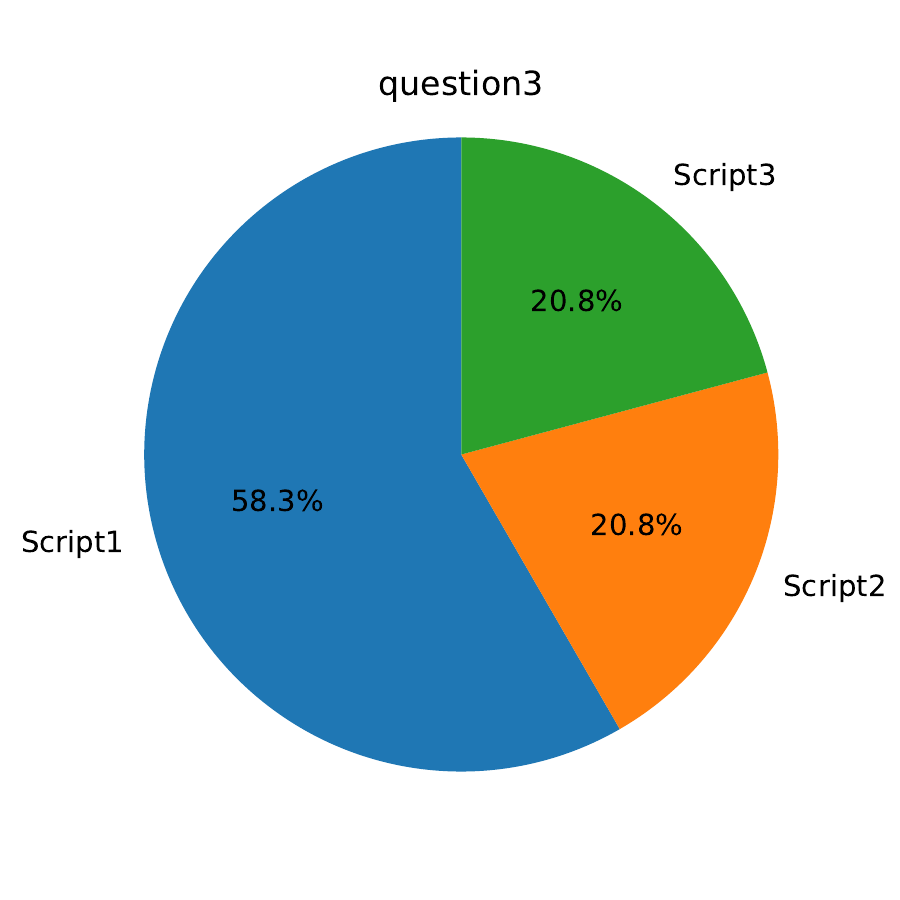}
		\caption*{Question 3}
	\end{minipage}
	\begin{minipage}[t]{0.19\textwidth}
		\centering
		\includegraphics[width=\linewidth]{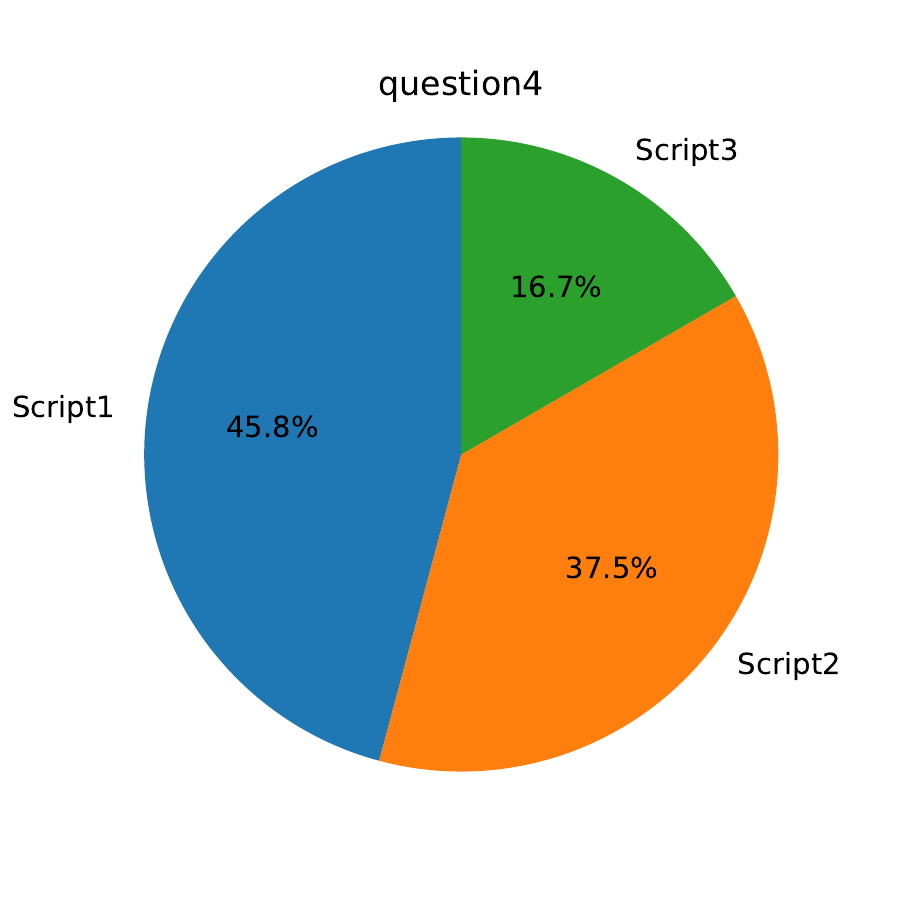}
		\caption*{Question 4}
	\end{minipage}
	\begin{minipage}[t]{0.19\textwidth}
		\centering
		\includegraphics[width=\linewidth]{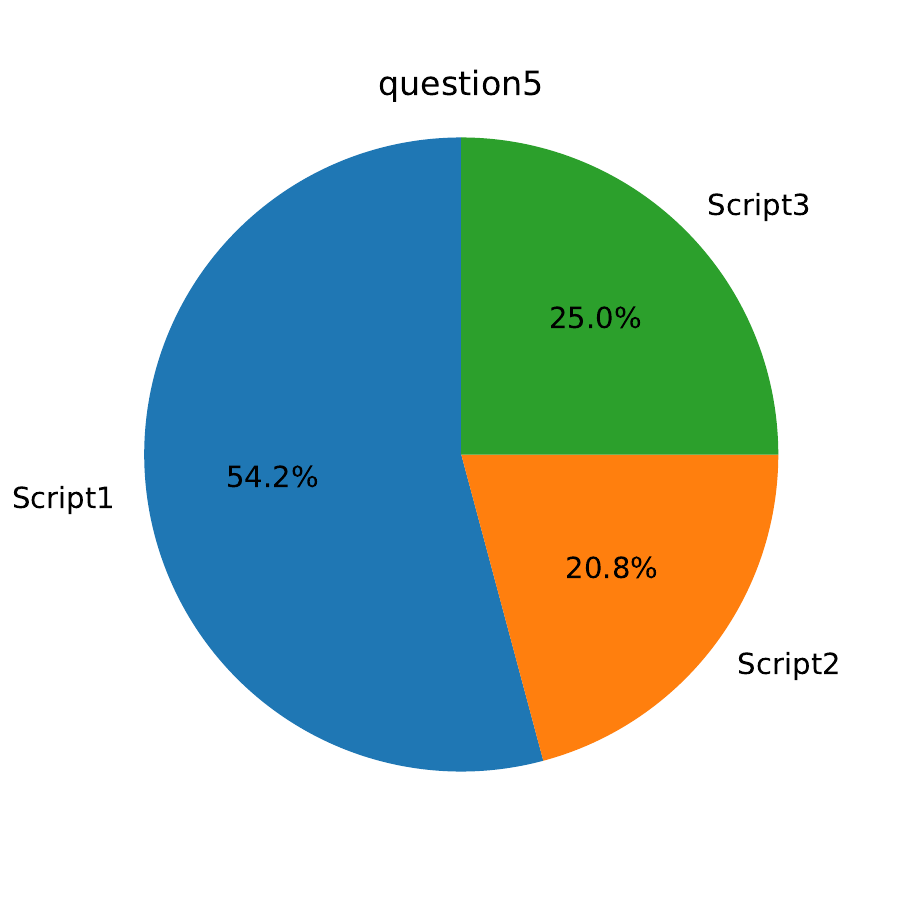}
		\caption*{Question 5}
	\end{minipage}
\end{figure}

\section*{Appendix F: Travel Itinerary Planning}
This section compares the proposed NarrativeGuide with several representative LLMs, including GPT-4o, GPT-4o-mini, and Qwen2.5-max, focusing on their travel itinerary planning capabilities. The comparison primarily examines travel time and the quality of the planned attractions, as shown in Tables \ref{tab:com_time} and \ref{tab:score}. Note that the base LLM used in NarrativeGuide is GPT-4o.

\begin{table}[ht]
	\scriptsize
	\centering
	\begin{tabular}{lcccc}
		\toprule
		& NarrativeGuide    & GPT-4o-mini & GPT-4o & Qwen2.5-max \\ \midrule  
		Nanjing  & \textbf{0.384} & 1.659      & 1.935  &1.854\\  \midrule
		Yangzhou & \textbf{0.786} & 2.656      & 1.645  &1.288\\  \midrule
		Paris    & \textbf{0.249} & 3.283      & 3.143  &3.566\\  \midrule
		Berlin   & \textbf{0.212} & 3.836      & 5.886  &3.595\\  \bottomrule
	\end{tabular}
	\caption{The travel time (h) of NarrativeGuide with GPT-4o are compared with the baseline GPT-4o, GPT-4o-mini, and Qwen2.5-max.}
	\label{tab:com_time}
\end{table}

\begin{table}[ht]
	\scriptsize
	\centering
	\resizebox{0.5\textwidth}{!}{
		\begin{tabular}{lcccc}
			\toprule
			& NarrativeGuide     & GPT-4o-mini & GPT-4o & Qwen2.5-max\\  \midrule
			Nanjing  & \textbf{3.79E+05} & 1.65E+05     & 8.87E+04 & 9.63E+04 \\  \midrule
			Yangzhou & \textbf{4.24E+05} & 9.40E+04      & 1.28E+05 & 1.71E+05 \\  \midrule
			Paris    & \textbf{5.17E+04}  & 1.08E+04      & 9.55E+03  & 9.55E+03\\  \midrule
			Berlin   & \textbf{1.30E+04}  & 4.11E+03       & 3.25E+03 & 3.67E+03 \\ \bottomrule
	\end{tabular}}
	\caption{The attraction score of NarrativeGuide with GPT-4o are compared with the baseline GPT-4o, GPT-4o-mini, and Qwen2.5-max.}
	\label{tab:score}
\end{table}

From the results in Table \ref{tab:com_time}, it is evident that with the introduction of GA optimization, the algorithm tends to recommend attractions that are clustered together, significantly reducing travel time. For example, for the city of Berlin, the travel time for the itinerary recommended by GPT-4o is 27 times greater than that of the NarrativeGuide. These results once again highlight that pure LLMs lack the capability for itinerary planning and are unable to suggest a reasonable travel plan. Additionally, the results in Table \ref{tab:score} further support this conclusion. The itineraries generated by NarrativeGuide have higher attraction scores, indicating that they feature popular destinations. However, this advantage is not as significant as the travel time reduction, as LLMs possess enough internal knowledge to recommend popular attractions. Yet, due to the inability to collect real-time data from the real world, this outcome is based on prior knowledge rather than updated, accurate data. Overall, the use of GA as an external planner in NarrativeGuide proves to be beneficial, significantly enhancing the ability of LLMs to address real-world problems and meet practical demands.

\appendix
\clearpage

\end{document}